\documentclass[prodmode,acmjdiq]{acmsmall} % Aptara syntax

\usepackage[ruled]{algorithm2e}

\SetAlFnt{\small}
\SetAlCapFnt{\small}
\SetAlCapNameFnt{\small}
\SetAlCapHSkip{0pt}
\IncMargin{-\parindent}

\usepackage{url}
\usepackage{latexsym}
\usepackage{color}
\usepackage[usenames]{xcolor}

\usepackage{amssymb}
\usepackage{graphicx}
\usepackage{rotating}
\usepackage{multirow}
\usepackage{pdfsync}
\usepackage{url}

 \newif\ifdraft
\newcommand{\blue}[1]{\ifdraft{\textcolor{blue}{#1}}\else{\textcolor{black}{#1}}\fi}

\newcommand{\hidden}[1]{\vspace{0ex}}

\usepackage[applemac]{inputenc} %

\begin{document}

% Page heads
\markboth{D. Marcheggiani and F. Sebastiani}{Effects of Low-Quality Training Data on Information
  Extraction}

% Title portion
\title{On the Effects of Low-Quality Training Data \\ on Information
  Extraction from Clinical Reports}
\author{DIEGO MARCHEGGIANI
\affil{Consiglio Nazionale delle Ricerche}
FABRIZIO SEBASTIANI
\affil{Qatar Computing Research Institute}
}
% NOTE! Affiliations placed here should be for the institution where the
%       BULK of the research was done. If the author has gone to a new
%       institution, before publication, the (above) affiliation should NOT be changed.
%       The authors 'current' address may be given in the "Author's addresses:" block (below).
%       So for example, Mr. Abdelzaher, the bulk of the research was done at UIUC, and he is
%       currently affiliated with NASA.

\begin{abstract}
In the last five years there has been a flurry of work on
   information extraction from clinical documents, i.e., on algorithms
   capable of extracting, from the informal and unstructured texts
   that are generated during everyday clinical practice, mentions of
   concepts relevant to such practice\footnotetext{Fabrizio Sebastiani is currently on leave from 
   Consiglio Nazionale delle Ricerche, Italy.}. Most of this literature is
   about methods based on supervised learning, i.e., methods for
   training an information extraction system from manually annotated
   examples. While a lot of work has been devoted to devising learning
   methods that generate more and more accurate information
   extractors, no work has been devoted to investigating the effect of
   the quality of training data on the learning process. Low quality
   in training data often derives from the fact that the person who
   has annotated the data is different from the one against whose
   judgment the automatically annotated data must be evaluated. In
   this paper we test the impact of such data quality issues on the
   accuracy of information extraction systems as applied to the
   clinical domain. We do this by comparing the accuracy deriving from
   training data annotated by the authoritative coder (i.e., the one
   who has also annotated the test data, and by whose judgment we must
   abide), with the accuracy deriving from training data annotated by
   a different coder. \blue{The results indicate that, although the
   disagreement between the two coders (as measured on the training
   set) is substantial, the difference is (surprisingly enough)
   not always statistically significant.}
\end{abstract}

\category{Information systems}{Information retrieval}{Retrieval tasks
and goals}[Clustering and Classification] \category{Computing
methodologies}{Machine learning}{Learning paradigms}[Supervised
learning]

\terms{Algorithm, Design, Experimentation, Measurements}


\acmformat{Diego Marcheggiani and Fabrizio Sebastiani, 2015. On the Effects of Low-Quality Training Data on Information Extraction from Clinical Reports.}
% At a minimum you need to supply the author names, year and a title.
% IMPORTANT:
% Full first names whenever they are known, surname last, followed by a period.
% In the case of two authors, 'and' is placed between them.
% In the case of three or more authors, the serial comma is used, that is, all author names
% except the last one but including the penultimate author's name are followed by a comma,
% and then 'and' is placed before the final author's name.
% If only first and middle initials are known, then each initial
% is followed by a period and they are separated by a space.
% The remaining information (journal title, volume, article number, date, etc.) is 'auto-generated'.

\begin{bottomstuff}
Author's addresses: Diego Marcheggiani, Istituto di Scienza e Tecnologie dell'Informazione,
Consiglio Nazionale delle Ricerche, 56124 Pisa, Italy; Fabrizio Sebastiani, Qatar Computing Research Institute, Qatar Foundation, PO Box 5825, Doha, Qatar. Fabrizio Sebastiani is currently on leave from Consiglio Nazionale delle Ricerche, Italy.   The order in which the authors are listed is purely alphabetical; each author has given an equally important contribution to this work.
\end{bottomstuff}

\maketitle

% ---------------------------------------------------------------------

\section{Introduction}\label{sec:introduction}

\noindent
% \fabscomment{Recommended reviewers: Ozlem Uzuner, Siddharta
% Jonnalagadda, Berry de Bruijn.} \fabscomment{Submit to the IEEE
% Transactions on Information Technology in Biomedicine; o anche al
% JAMIA.}
In the last five years there has been a flurry of work (see e.g.,
\cite{kelly2014overview,Pradhan2014,Sun:2013yq,Suominen:2013eu,Uzuner:2012fp,Uzuner2011})
on information extraction from clinical documents, i.e., on algorithms
capable of extracting, from the informal and unstructured texts that
are generated during everyday clinical practice (e.g., admission
reports, radiological reports, discharge summaries, clinical notes),
mentions of concepts relevant to such practice. Most of this
literature is about methods based on supervised learning, i.e.,
methods for training an information extraction system from manually
annotated examples. While a lot of work has been devoted to devising
text representation methods and variants of the aforementioned
supervised learning methods that generate more and more accurate
information extractors, no work has been devoted to investigating the
effects of the quality of training data on the learning
process. Issues of quality in the training data may arise for
different reasons:

\begin{enumerate}

\item \label{ref:fast} In several organizations it is often the case
  that the original annotation is performed by coders (a.k.a.\
  ``annotators'', or ``assessors'') as a part of a daily routine in
  which fast turnaround, rather than annotation quality, is the main
  goal of the coders and/or of the organization. An example is the
  (increasingly frequent) case in which annotation is performed via
  crowdsourcing using instruments such as, e.g., Mechanical Turk,
  CrowdFlower, etc.\footnote{\url{https://www.mturk.com/},
  \url{http://crowdflower.com/}} \cite{Grady:172vn,Snow:2008kx}.
 
\item In many organizations it is also the case that annotation work
  is usually carried out by junior staff (e.g., interns), since having
  it accomplished by senior employees would make costs soar.
 
\item It is often the case that the coders entrusted with the
  annotation work were not originally involved in designing the tagset
  (i.e., the set of concepts whose mentions are sought in the
  documents). As a result, the coders may have a suboptimal
  understanding of the true meaning of these concepts, or of how their
  mentions are meant to look like, which may negatively affect the
  quality of their annotation.
 
\item \label{ref:old} The data used for training the system may
  sometimes be old or outdated, with the annotations not reflecting
  the current meaning of the concepts anymore. This is an example of a
  phenomenon, called \emph{concept drift}
  \cite{Quinonero09,Sammut:2011fk}, which is well known in machine
  learning.
 
\end{enumerate}
 
\noindent We may summarize all the cases mentioned above by saying
that, should the training data be independently re-annotated by an
\emph{authoritative coder} (hereafter indicated as $C_{\alpha}$), the
resulting annotations would be, to a certain extent, more reliable. We
would also be able to precisely measure this difference in
reliability, by measuring the \emph{intercoder agreement}
% $IA_{Tr}(C_{\alpha},C_{\beta})$
(via measures such as Cohen's kappa -- see e.g.,
\cite{Artstein:2008kx,DiEugenio04}) between the training data $Tr$ as
coded by $C_{\alpha}$ and the training data as coded by whoever else
originally annotated them (whom we will call, for simplicity, the
\emph{non-authoritative coder} -- hereafter indicated as $C_{\beta}$). In
the rest of this paper we will take the authoritative coder
$C_{\alpha}$ to be \emph{the coder whose annotations are to be taken
as correct}, i.e., considered as the ``gold standard''. As a
consequence we may assume that $C_{\alpha}$ is the coder who, once the
system is trained and deployed, has also the authority to evaluate the
accuracy of the automatic annotation (i.e., decide which annotations
are correct and which are not)\footnote{In some organizations this
authoritative coder may well be a fictional entity, e.g., several
coders may be equally experienced and thus equally
authoritative. However, without loss of generality we will hereafter
assume that $C_{\alpha}$ exists and is unique.}. In this case,
intercoder (dis)agreement
% $ICA_{Tr}(C_{\alpha},C_{\beta})$ (or rather its complement:
% -$ICA_{Tr}(C_{\alpha},C_{\beta})$)
measures the amount of noise that is introduced in the training data
by having them annotated by a coder $C_{\beta}$ different from the
authoritative coder $C_{\alpha}$.

% \fabscomment{Could we call this $Tr_{\alpha}$-disagreement?}

It is natural to expect the accuracy of an information extraction
system to be lower if the training data have been annotated by a
non-authoritative coder $C_{\beta}$, and higher if they have been annotated
by $C_{\alpha}$ herself. However, note that this is \emph{not} a
consequence of the fact that $C_{\alpha}$ is more experienced, or
senior, or reliable, than $C_{\beta}$. Rather, it is a consequence of
the fact that standard supervised learning algorithms are based on the
assumption that the training set and the test set are identically and
independently distributed (the so-called \emph{i.i.d.\ assumption}),
i.e., that both sets are randomly drawn from the \emph{same}
distribution. As a result, these algorithms learn to replicate the
subjective annotation style of their supervisors, i.e., of those who
have annotated the training data. This means that we may expect
accuracy to be higher simply when the coder of the training set and
the coder of the test set are the \emph{same} person, and to be lower
when the two coders are different, irrespective of how experienced, or
senior, or reliable, they are. In other words, the very fact that a
coder is entrusted with the task of evaluating the automatic
annotations (i.e., of annotating the test set) makes this coder
\emph{authoritative by definition}. For this reason, the authoritative
coder $C_{\alpha}$ may equivalently be defined as ``the coder who has
annotated the test set'' (or: ``the coder whose judgments we adhere to
when evaluating the accuracy of the system''), and the non-authoritative
coder $C_{\beta}$ may equivalently be defined as ``a coder different
from the authoritative coder''.

The above arguments point to the fact that the impact of training data
quality -- under its many facets discussed in items
(\ref{ref:fast})-(\ref{ref:old}) above -- on the accuracy of information
extraction systems may be measured by

\begin{enumerate}

\item evaluating the accuracy of the system in an \emph{authoritative}
  setting (i.e., both training and test sets annotated by the
  authoritative coder $C_{\alpha}$), and then
 
\item evaluating the loss in accuracy, with respect to the authoritative
  setting, that derives from working instead in a \emph{non-authoritative}
  setting (i.e., test set annotated by $C_{\alpha}$ and training set
  annotated by a non-authoritative coder $C_{\beta}$)\footnote{In the
  domain of classification the authoritative and non-authoritative settings
  have also been called \emph{self-classification} and
  \emph{cross-classification}, respectively \cite{Webber:2013fk}. We
  depart from this terminology in order to avoid any confusion with
  \emph{self-learning} (which refers to retraining a classifier by
  using, as additional training examples, examples the classifier
  itself has classified) and \emph{cross-lingual classification}
  (which denotes a variant of text classification which exploits
  synergies between training data expressed in different languages).}.
 
\end{enumerate}

% ---------------------------------------------------------------------

\subsection{Our contribution}\label{sec:ourcontribution}

\noindent In this paper we test the impact of training data quality on
the accuracy of information extraction systems as applied to the
clinical domain. We do this by testing the accuracy of two state-of
the-art systems on a dataset of radiology reports (originally
discussed in \cite{Esuli:2013uq}) in which a portion of the data has
independently been annotated by two different experts. In other words,
we try to answer the question: ``What is the consequence of the fact
that my training data are not sterling quality? that the coders who
produced them are not entirely dependable? How much am I going to lose
in terms of accuracy of the trained system?''

In these experiments we not only test the ``pure'' authoritative and
non-authoritative settings described above, but we also test
\emph{partially authoritative} settings, in which increasingly large
portions of the training data as annotated by $C_{\alpha}$ are
replaced with the corresponding portions as annotated by $C_{\beta}$,
thus simulating the presence of incrementally higher amounts of
noise. For each setting we compute the intercoder agreement between
the two training sets; this allows us to study the relative loss in
extraction accuracy as a function of the agreement between
authoritative and non-authoritative assessor as measured on the training
set. Since in many practical situations it is easy to compute (or
estimate) the intercoder disagreement between (a) the coder to whom we
would ideally entrust the annotation task (e.g., a senior expert in
the organization), and (b) the coder to whom we can indeed entrust it
given time and cost constraints (e.g., a junior member of staff), this
will give the reader a sense of how much intercoder disagreement
generates how much loss in extraction accuracy.

The rest of the paper is organized as follows. Section
\ref{sec:relatedwork} reviews related work on information extraction
from clinical documents, and on establishing the relations between
training data quality and extraction accuracy. In Sections
\ref{sec:experiments} and \ref{sec:results} we describe experiments
that attempt to quantify the degradation in extraction accuracy that
derives from low-quality training data, with Section
\ref{sec:experiments} devoted to spelling out the experimental setting
and Section \ref{sec:results} devoted instead to presenting and
discussing the results. Section \ref{sec:conclusions} concludes,
discussing avenues for further research.

% ---------------------------------------------------------------------

\section{Related Work}\label{sec:relatedwork}

% ---------------------------------------------------------------------

\subsection{Information extraction from clinical
documents}\label{sec:IE}

\noindent Most works on information extraction from clinical documents
rely on methods based on supervised learning, i.e., methods for
training an information extraction system from manually annotated
examples. Support vector machines (SVMs --
\cite{Jiang:2011ve,Li:2008uq,Sibanda:2006fk}), hidden Markov models
(HMMs -- \cite{Li:2010uq}), and (especially) conditional random fields
(CRFs --
\cite{Esuli:2013uq,Gupta:2014yq,Jiang:2011ve,Jonnalagadda:2012uq,Li:2008uq,Patrick:2010fk,Torii:2011ly,Wang:2009fk})
have been the learners of choice in this
field, 
due to their good performance and to the existence of publicly
available implementations.
%\blue{Many such works have taken a fairly domain-independent
%approach to clinical information extraction, i.e., one that exploits
%domain-independent information extraction techniques (e.g., use of
%sequence learning algorithms, use of a generic NLP pipeline,
%geared to the clinical domain. In contrast, other works (see e.g.,
%\cite{Friedman:2002qd}) have tried (following Zellig Harris' theory of
%sub-languages \cite{Harris2002}) to exploit the features of the
%``clinical sub-language'', often in the form of specific vocabulary, specialized ontologies, 
%specialized semantic relations, or even specialized syntax. The present work belongs to the
%former camp.}

\blue{In recent years, research on the analysis of clinical texts has
been further boosted by the existence of ``shared tasks'' on this
topic, such as the seminal i2b2 series (``Informatics for Integrating
Biology and the Bedside'' --
\cite{Sun:2013yq,Uzuner:2012fp,Uzuner2011}), the 2013
\cite{Suominen:2013eu} and 2014 \cite{kelly2014overview} editions of
ShARe/CLEF eHealth, and the Semeval-2014 Task 7 ``Analysis of Clinical
Text'' \cite{Pradhan2014}. In these shared tasks the goal is to
competitively evaluate information extraction tools that recognise
mentions of various concepts of interest (e.g., mentions of diseases
and disorders) as appearing in discharge summaries, and in
electrocardiogram reports, echocardiograph reports, and radiology
reports.}

% \diescomment{ The learning methods the participants have adopted for
% these tasks are mostly CRFs and HM-SVMs. What is new in the
% approaches proposed by most of the participants, e.g.,
% \cite{attardi2014unipi,Zhang2014}, are the features and the
% variations of the IOB format formulations in order to cope with the
% discontinuous mentions present in the dataset. These aspects are
% invariants of the problem we study in this work, i.e., the effects
% of the quality of labels of the training data on the learning
% process. }

% \fabscomment{Mention the fact that the dataset described in
% \cite{Uzuner:2010fk} contains extensive intercoder data; I am not
% really sure whether anyone has used the data, though.}

% ---------------------------------------------------------------------

\subsection{Low-quality training data and prediction
accuracy}\label{sec:lowquality}

\noindent The literature on the effects of suboptimal training data
quality on prediction accuracy is extremely scarce, even within the
machine learning literature at large. An early such study is
\cite{Rossin:1999uq}, which looks at these issues in the context of
learning to predict prices of mutual funds from economic
indicators. Differently from us, the authors work with noise
artificially inserted in the training set, and not with naturally
occurring noise. From experiments run with a linear regression model
they reach the bizarre conclusion that ``the predictive accuracy (...)
is better when errors exist in training data than when training data
are free of errors'', while the opposite conclusion is (somehow more
expectedly) reached from experiments run with a neural networks
model. A similar study, in which the context is predicting the average
air temperature in distributed heating systems, was carried out in
\cite{Jassar:2009kx}; yet another study, in which the goal was
predicting the production levels of palm oil via a neural network, is
\cite{Khamis:2005vn}.

In the context of a biomedical information extraction
task\footnote{Biomedical IE is different from clinical IE, in that the
latter (unlike the former) is usually characterized by idiosyncratic
abbreviations, ungrammatical sentences, and sloppy language in
general. See \cite[p.\ 129]{Meystre:2008vn} for a discussion of this
point.} Haddow and Alex \citeyear{Haddow:2008fk} examined the situation in
which training data annotated by two different coders are available,
and they found that higher accuracy is obtained by using both versions
at the same time than by attempting to reconcile them or using just
one of them. Their use case is different from ours, since in the case
we discuss we assume that only one set of annotations, those of the
non-authoritative coder, are available as training data. Note also that
training data independently annotated by more than one coder are
rarely available in practice.

Closer to our application context, Esuli and Sebastiani
\citeyear{Esuli:2013ko} have thoroughly studied the effect of suboptimal
training data quality in text classification. However, in their case
the degradation in the quality of the training data is obtained, for
mere experimental purposes, via the insertion of artificial noise, due
to the fact that their datasets did not contain data annotated by more
than one coder. As a result, it is not clear how well the type of
noise they introduce models naturally occurring noise. Webber and
Pickens \citeyear{Webber:2013fk} also address the text classification task
(in the context of e-discovery from legal texts), but differently from
\cite{Esuli:2013ko} they work with naturally occurring noise;
differently from the present work, the multiply-coded training data
they use were coded by one coder known to be an expert coder and
another coder known to be a junior coder. Our work instead (a) focuses
on information extraction, and (2) does not make any assumption on the
relative level of expertise of the two coders.

% \fabscomment{William Webber has two interesting blog
% posts\footnote{William Webber, \emph{Annotator error and predictive
% reliability}, 27 Dec 2013, \url{http://blog.codalism.com/?p=2060};
% William Webber, \emph{Assessor error and term model weights}, 3 Jan
% 2014, \url{http://blog.codalism.com/?p=2068}.} on this problem,
% which contain references to other pieces of research who have
% investigated the problem. See also \cite{Scholtes:2013fk}.}
%
% \fabscomment{\cite{Li:2014fk} tests the sensitivity of an evaluation
% measure to relevance judgments expressed by a non-authoritative
% assessor.}

% \ronescomment{The paper addresses the issue of how the accuracy of
% training data affects the accuracy of supervised training algorithms
% for the analysis of clinical documents. The paper should mention the
% shared tasks that have been organized in recent years on the
% analysis of clinical texts, including Semeval-2014 Task 7,
% ShARe/CLEF eHealth 2013 \cite{Pradhan2014,Suominen:2013eu}. Some of
% the best approaches combine supervised methods with unsupervised
% methods exploiting distributional semantics, in the form of word
% embeddings, random indexing or word clusters. Unsupervised methods
% may be helpful to reduce reliance on manually annotated
% datasets. See, for example \cite{Zhang2014}.}

% ---------------------------------------------------------------------

\section{Experimental setting}\label{sec:experiments}

% ---------------------------------------------------------------------

\subsection{Basic notation and terminology}\label{sec:notation}

\noindent Let us fix some basic notation and terminology. Let
$\mathbf{X}$ be a set of texts, where \blue{we view each text
$\mathbf{x}\in \mathbf{X}$ as a sequence $\mathbf{x}=\langle x_{1},
\ldots, x_{|\mathbf{x}|}\rangle$ of \emph{textual units} (or simply
\emph{t-units}), such that odd-numbered t-units are \emph{tokens}
(i.e., word occurrences) and even-numbered t-units are
\emph{separators} (i.e., sequences of blanks and punctuation
symbols)}, and such that $x_{t_{1}}$ occurs before $x_{t_{2}}$ in the
text (noted $x_{t_{1}}\preceq x_{t_{2}}$) if and only if $t_{1}\leq
t_{2}$. We dub $|\mathbf{x}|$ the \emph{length} of the text. Let
$C=\{c_{1}, \ldots, c_{m}\}$ be a predefined set of \emph{concepts}
(a.k.a.\ \emph{tags}, or \emph{markables}), or \emph{tagset}. We take
\emph{information extraction} (IE) to be the task of determining, for
each $\mathbf{x}\in \mathbf{X}$ and for each $c_{r}\in C$, a sequence
$\mathbf{y}_{r}=\langle y_{r1}, \ldots, y_{r|\mathbf{x}|}\rangle$ of
\emph{labels} $y_{rt}\in\{c_{r},\overline{c}_{r}\}$, which indicates
which \blue{t-units} in the text are labelled with tag $c_{r}$ and
which are not. Since each $c_{r}\in C$ is dealt with independently of
the other concepts in $C$, we hereafter drop the $r$ subscript and,
without loss of generality, treat IE as the \emph{binary} task of
determining, given text $\mathbf{x}$ and concept $c$, a sequence
$\mathbf{y}=\langle y_{1}, \ldots, y_{|\mathbf{x}|}\rangle$ of labels
$y_{t}\in\{c,\overline{c}\}$.

\blue{T-units} labelled with a concept $c$ usually come in coherent
sequences, or ``mentions''. Hereafter, a \emph{mention} $\sigma$ of
text $\mathbf{x}$ for concept $c$ will be a pair
$(x_{t_{1}},x_{t_{2}})$ consisting of a start token $x_{t_{1}}$ and an
end token $x_{t_{2}}$ such that (i) $x_{t_{1}} \preceq x_{t_{2}}$,
(ii) all \blue{t-units} $x_{t_{1}} \preceq x_{t} \preceq x_{t_{2}}$
are labelled with concept $c$, and (iii) the token that immediately
precedes $x_{t_{1}}$ and the one that immediately follows $x_{t_{2}}$
are \emph{not} labelled with concept $c$. In general, a text
$\mathbf{x}$ may contain zero, one, or several mentions for concept
$c$.

\blue{In the above definitions we consider separators to be also the
object of tagging in order for the IE system to correctly identify
consecutive mentions. For instance, given the expression ``Barack Obama,
Hillary Clinton'' the perfect IE system will attribute the
\textsf{PersonName} tag to the tokens ``Barack'', ``Obama'',
``Hillary'', ``Clinton'', and to the separators (in this case: blank
spaces) between ``Barack'' and ``Obama'' and between ``Hillary'' and
``Clinton'', but \emph{not} to the separator ``, '' between ``Obama''
and ``Hillary''. If the IE system does so, this means that it has
correctly identified the boundaries of the two mentions ``Barack
Obama'' and ``Hillary Clinton''\footnote{\blue{Note that the above
notation is not able to represent ``discontiguous mentions'', i.e.,
mentions containing gaps, and ``overlapping mentions'', i.e., multiple
mentions sharing one or more tokens. This is not a serious limitation
for our research, since the above notation can be easily extended to
deal with both phenomena (e.g., by introducing unique mention
identifiers and having each t-unit be associated with zero, one, or
several such identifiers), and since the dataset we use for our
experimentation contains neither discontinuous nor overlapping
mentions. We prefer to keep the notation simple, since the issue we
focus on in this paper (the consequences on extraction accuracy of
suboptimal training data quality) can be considered largely
independent of the expressive power of the markup language.}}.}

\subsection{Dataset}\label{sec:dataset}

\noindent
% \fabscomment{Dovremmo investigare se ci sono altri dataset per
% information extraction from clinical text che abbiano dati di
% intercoder agreement. E.g., i dataset i2b2, il dataset CLEF, il
% dataset di \cite{Roberts:2009fk}. Guardare nella ``related work
% section'' di \cite{Roberts:2009fk}.}
The dataset we have used to test the ideas discussed in the previous
sections is the UmbertoI(RadRep) dataset first presented in
\cite{Esuli:2013uq}, consisting of a set of 500 free-text mammography
reports written (in Italian) by medical personnel of the Istituto di
Radiologia of Policlinico Umberto I, Roma, IT. The dataset is
annotated according to 9 concepts relevant to the field of radiology
and mammography: BIR (``Outcome of the BIRADS test''), ITE
(``Technical Info''), IES (``Indications obtained from the Exam''),
TFU (``Followup Therapies''), DEE (``Description of Enhancement''),
PAE (``Presence/Absence of Enhancements''), ECH (``Outcomes of
Surgery''), DEP (``Prosthesis Description''), and LLO (``Locoregional
Lymph Nodes''). \blue{Note that we had no control on the design of the
concept set, on its range, and on its granularity, since the choice of
the concepts was entirely under the responsibility of Policlinico
Umberto I. We thus take both the concept set and the dataset as
given.}

Mentions of these concepts are present in the reports according to
fairly irregular patterns. In particular, a given concept (a) need not
be instantiated in all reports, and (b) may be instantiated more than
once (i.e., by more than one mention) in the same report. Mentions
instantiating different concepts may overlap, and the order of
presentation of the different concepts varies across the reports. On
average, there are 0.87 mentions for each concept in a given report,
and the average mention length is 17.33 words.

The reports were annotated by two equally expert radiologists, Coder1
and Coder2; 191 reports were annotated by Coder1 only, 190 reports
were annotated by Coder2 only, and 119 reports were annotated
independently by Coder1 and Coder2. From now on we will call these
sets \texttt{1-only}, \texttt{2-only} and \texttt{Both}, respectively;
\texttt{Both(1)} will identify the \texttt{Both} set as annotated by
Coder1, and \texttt{Both(2)} will identify the \texttt{Both} set as
annotated by Coder2. The annotation activity was preceded by an
alignment phase, in which Coder1 and Coder2 jointly annotated 20
reports (not included in this dataset) in order to align their
understanding of the meaning of the concepts.

\blue{Table \ref{tab:annotationstats} reports the distribution of
annotations across concepts, at token and mention level, for the two
coders;} see \cite[Section 4.2]{Esuli:2013uq} for a more detailed
description of the UmbertoI(RadRep) dataset that includes additional
stats\footnote{\label{foot:datasets}No other dataset is used in this
paper since we were not able to locate another dataset of annotated
clinical texts that contains reports annotated by more than one coder
and is at the same time publicly available.}.

\begin{table}[t]
  \begin{center}
    \tbl{\label{tab:annotationstats}\blue{The distribution of annotations across concepts, at token and mention level, for each coder.}}{
    \resizebox{\textwidth}{!} {
    \begin{tabular}{|r||r|r|r|r|r|r|r|r|r||r|}
      \hline
      & DEE & IES & ITE & ECH & LLO & TFU & DEP & BIR & PAE & Total\\
      \hline\hline
      Tokens annotated by Coder1 & 4819 & 1529 & 7410 & 237 & 1811 & 1672 & 585 & 466 & 1723 & 18529
      \\\hline
      Tokens annotated by Coder2 & 7351 & 1723 & 7630 & 1329 & 2544 & 2670 & 1127 & 448 & 3495 & 24822
      \\
      \hline
      Mentions annotated by Coder1 & 204 & 140 & 190 & 51 & 164 & 149 & 19 & 128 & 344 & 1045
      \\\hline
      Mentions annotated by Coder2 & 282 & 145 & 188 & 102 & 193 & 171 & 26 & 103 & 399 & 1210
      \\
      \hline
    \end{tabular}
    }}
  \end{center}
\end{table}

% \fabscomment{Si possono citare quali dataset non contengono report
% annotati da più di un annotatore e quali non sono invece
% pubblicamente disponibili (\cite{Chapman:2006kx}). The dataset
% discussed in \cite{Roberts:2009fk} is not publicly
% available. \fabscomment{\cite{Uzuner:2010fk} reports on an
% experiment of collective annotation of discharge summaries by the
% participants in the i2b2 task, and describes a dataset that contains
% extensive intercoder data; \cite{Uzuner:2010uq} is an overview of
% the results obtained by participants in the task on this dataset; we
% should probably check if any of the participants has indeed made use
% of the intercoder agreement data in their systems.}.

% ---------------------------------------------------------------------

\subsection{Learning algorithms}\label{sec:learner}

\noindent As the learning algorithms we have tested both
\emph{linear-chain conditional random fields} (LC-CRFs -
\cite{Lafferty01,Sutton:2007uq,Sutton:2012fk}), in Charles Sutton's
GRMM implementation\footnote{\url{http://mallet.cs.umass.edu/grmm/}},
\blue{and \emph{hidden Markov support vector machines} (HM-SVMs -
\cite{Altun03hidden}), in Thorsten Joachims's $SVM^{hmm}$
implementation\footnote{\url{http://www.cs.cornell.edu/people/tj/svm_light/svm_hmm.html}}.}
Both are supervised learning algorithms explicitly devised for
\emph{sequence labelling}, i.e., for learning to label (i.e., to
annotate) items that naturally occur in sequences and such that the
label of an item may depend on the features and/or on the labels of
other items that precede or follow it in the sequence (which is indeed
the case for the tokens in a text)\footnote{\blue{\label{foot:separators}Note that only tokens, and not separators, are explicitly labelled. The reason is that both LC-CRFs and HM-SVMs actually use the so-called IOB labelling scheme, according to which, for each concept $c_{r}\in C$, a token can be labelled as $B_{r}$ (the \underline{b}eginning token of a mention of $c_{r}$), $I_{r}$ (a token which is \underline{i}nside a mention of $c_{r}$ but is not its beginning token), and $O_{r}$ (a token that is \underline{o}utside any mention of $c_{r}$). As a result, a separator is (implicitly) labelled with concept $c_{r}$ if and only if it precedes a token labelled with $I_{r}$. We may think of the notation of Section \ref{sec:notation} as an abstract markup language, and of the IOB notation as a concrete markup language, in the sense that the notation of Section \ref{sec:notation} is easier to understand (and will also make the evaluation measure discussed in Section \ref{sec:classificationaccuracy} easier to understand) while IOB is actually used by the learning algorithms. The two notations are equivalent in expressive power.}}. LC-CRFs are members of the class
of \emph{graphical models}, a family of probability distributions that
factorize according to an underlying graph \cite{Wainwright:2008fk};
see \cite{Sutton:2012fk} for a full mathematical explanation of
LC-CRFs. \blue{HM-SVMs are an instantiation of ``SVMs for structured
output prediction'' ($SVM^{struct}$) \cite{tsochantaridis05large} for
the sequence labelling task, and have already been used in clinical
information extraction (see e.g.,
\cite{tang2012clinical,Zhang2014}). In HM-SVMs the learning procedure
is based on a large-margin approach typical of SVMs, which,
differently from LC-CRFs, can learn non-linear discriminant functions
via kernel functions.}

Both learners need each token $x_{t}$ to be represented by a vector
$\mathbf{x}_{t}$ of features\footnote{\blue{Note that only tokens, and not separators, are explicitly represented in vectorial form, the reasons being the same already discussed in Footnote \ref{foot:separators}.}}. In this work we have used a set of
features which includes one feature representing the word of which the
token is an instance, one feature representing its stem, one feature
representing its part of speech, eight features representing its
prefixes and suffixes (the first and the last $n$ characters of the
token, with $n=1, 2, 3, 4$), one feature representing information on
token capitalization (i.e., whether the token is all uppercase, all
lowercase, first letter uppercase, or mixed case), and 4
``positional'' features \cite[Section 3.3]{Esuli:2013uq} that indicate
in which half, 3rd, 4th, or 5th, respectively, of the text the token
occurs in. 
%\fabscomment{Diego, i separators invece non vengono
%rappresentati da feature vectors?}

% \fabscomment{Sarebbe bello mettere a disposizione i vettori che
% abbiamo generato in modo da permettere ad altri di replicare i
% nostri risultati. Ma se il dataset non è disponibile ciò non è
% possibile.}

% ---------------------------------------------------------------------

\subsection{Evaluation measures}\label{sec:evaluationmeasure}

% ---------------------------------------------------------------------

\subsubsection{Classification
accuracy}\label{sec:classificationaccuracy}

\noindent As a measure of classification accuracy we use, similarly to
\cite{Esuli:2013uq}, the token-and-separator variant (proposed
in~\cite{Esuli:2010jx}) of the well-known $F_{1}$ measure, according
to which an information extraction system is evaluated on an event
space consisting of all the \blue{t-units} in the text. In other words, each
\blue{t-unit} $x_{t}$ (rather than each \emph{mention}, as in the traditional
``segmentation F-score'' model \cite{Suzuki:2006fk}) counts as a true
positive, true negative, false positive, or false negative for a given
concept $c_{r}$, depending on whether $x_{t}$ belongs to $c_{r}$ or
not in the predicted annotation and in the true annotation. This model
has the advantage that it credits a system for partial success (i.e.,
degree of overlap between a predicted mention and a true mention for
the same concept), and that it penalizes both overannotation and
underannotation.

As is well-known, $F_{1}$ \blue{is the harmonic mean of}
\emph{precision} ($\pi=\frac{TP}{TP+FN}$) and \emph{recall}
($\rho=\frac{TP}{TP+FP}$), and is defined as
\begin{equation}
  \label{eq:F1}F_{1}=\displaystyle\frac{2 \pi \rho}{\pi + \rho}\blue{=\displaystyle\frac{2\cdot \displaystyle\frac{TP}{TP+FN}\cdot \displaystyle\frac{TP}{TP+FP}}{\displaystyle\frac{TP}{TP+FN}+\displaystyle\frac{TP}{TP+FP}}}=\displaystyle\frac{2 TP}{2 TP+FP+FN}
\end{equation} 
\noindent where $TP$, $FP$, and $FN$ stand for the numbers of true
positives, false positives, and false negatives, respectively.
\blue{It is easy to observe that $F_{1}$ is equivalent to $TP$ divided
by the arithmetic mean of the actual positives and the predicted
positives (or, alternatively, the product of $\pi$ and $\rho$ divided
by their arithmetic mean).} Note that $F_{1}$ is undefined when
$TP=FP=FN=0$; in this case we take $F_{1}$ to equal $1$, since the
system has correctly annotated all \blue{t-units} as negative.

We compute $F_{1}$ across the entire test set, i.e., we generate a
single contingency table by putting together all \blue{t-units} in the test
set, irrespectively of the document they belong to. We then compute
both \emph{microaveraged} $F_{1}$ (denoted by $F_{1}^{\mu}$) and
\emph{macroaveraged} $F_{1}$ ($F_{1}^{M}$). $F_{1}^{\mu}$ is obtained
by (i) computing the concept-specific values $TP_{r}$, $FP_{r}$ and
$FN_{r}$, (ii) obtaining $TP$ as the sum of the $TP_{r}$'s (same for
$FP$ and $FN$), and then (iii) applying Equation \ref{eq:F1}.
$F_{1}^{M}$ is obtained by first computing the concept-specific
$F_{1}$ values and then averaging them across the $c_{r}$'s.

% ---------------------------------------------------------------------

\subsubsection{Intercoder agreement}\label{sec:intercoderagreement}

% An advantage of using $F_{1}$ as the evaluation measure is that it
% is symmetric, i.e., its values do not change if one switches the
% roles of the human coder (i.e., the gold standard) and the automatic
% coder (i.e., the system). This means that $F_{1}$ can also be used
% as a measure of agreement between any two coders, regardless of
% whether they are human or machine, since it does not require one to
% specify who among the two is the gold standard against which the
% other needs to be checked. For this reason, in the following section
% we will use $F_{1}$ both (a) to measure the agreement between our
% system and a human coder, \emph{and} (b) to measure the agreement
% between two human coders. This will allow us to judge in a direct
% way how our system fares with respect to human performance.

\noindent \emph{Intercoder agreement} (ICA), or the lack thereof
(\emph{intercoder disagreement}), has been widely studied for over a
century (see e.g., \cite{Krippendorff:2004kx} for an introduction). As
a phenomenon, disagreement among coders naturally occurs when units of
content need to be annotated by humans according to their semantics
(i.e., when the occurrences of specific concepts need to be recognized
within these units of content). Such disagreement derives from the
fact that semantic content is a highly subjective notion: different
coders might disagree with each other as to what the semantics of,
say, a given piece of text is, and it is even the case that the same
coder might at times disagree with herself (i.e., return different
codes when coding the same unit of content at different times).

ICA may be measured by the relative frequency of the units of content
on which coders agree, usually normalized by the probability of chance
agreement. Many metrics for ICA have been proposed over the years,
``Cohen's kappa'' probably being the most famous and widely used
(``Scott's pi'' and ``Krippendorff's alpha'' are others); sometimes
(see e.g., \cite{Chapman:2006kx,Esuli:2013uq}) functions that were not
explicitly developed for measuring ICA (such as $F_{1}$, that was
developed for measuring binary classification accuracy) are used. The
levels of ICA that are recorded in actual experiments vary a lot
across experiments, types of content, and types of concepts that are
to be recognized in the units of content under investigation. This
extreme variance depends on factors such as ``annotation domain,
number of categories in a coding scheme, number of annotators in a
project, whether annotators received training, the intensity of
annotator training, the annotation purpose, and the method used for
the calculation of percentage agreements'' \cite{Bayerl:2011uq}. The
actual meaning of the concepts the coders are asked to recognize is a
factor of special importance, to the extent that a concept on which
very low levels of ICA are reached may be deemed, because of this very
fact, ill-defined.

For measuring intercoder agreement we use Cohen's kappa (noted
$\kappa$), defined as
\begin{eqnarray}
  \label{eq:kappa}\kappa & = & \frac{P(A)-P(E)}{1-P(E)} \\
  & = & \blue{\frac{(P(p=t=c)+P(p=t=\overline{c}))-(P(p=c)P(t=c)+P(p=\overline{c})P(t=\overline{c}))}{1-(P(p=c)P(t=c)+P(p=\overline{c})P(t=\overline{c}))}} \nonumber \\
  & = & \blue{\frac{\displaystyle\frac{TP+TN}{n}-((\frac{TP+FP}{n})(\frac{TP+FN}{n})+(\frac{FN+TN}{n})(\frac{FP+TN}{n}))}{1-((\displaystyle\frac{TP+FP}{n})(\frac{TP+FN}{n})+(\frac{FN+TN}{n})(\frac{FP+TN}{n}))}} \nonumber 
\end{eqnarray} 
\noindent where $P(A)$ denotes the probability (i.e., relative
frequency) of agreement, $P(E)$ denotes the probability of chance
agreement, and $n$ is the total number of examples (see
\cite{Artstein:2008kx,DiEugenio04} for details)\blue{; here, we use
the shorthand $p=c$ (resp., $t=c$) to mean that the
\underline{p}redicted label (resp., \underline{t}rue label) is $c$
(analogously for $\overline{c}$)}. We opt for kappa since it is the
most widely known, and best understood, measure of ICA. For Cohen's
kappa too we work at the \blue{t-unit} level, i.e., for each \blue{t-unit} $x_{t}$
we record whether the two coders agree on whether $x_{t}$ is labelled
or not with the concept $c$ of interest.

Incidentally, note that (as observed in \cite{Esuli:2010jx}) we can
compute Cohen's kappa only thanks to the fact that (as discussed in
Section \ref{sec:classificationaccuracy}) we conduct our evaluation at
the \blue{t-unit} level (rather at the mention level). Those who conduct
their evaluation at the mention level (e.g., \cite{Chapman:2006kx})
find that they are unable to do so, since in order to be defined kappa
needs the notion of a true negative to be also defined, and this is
undefined at the mention level. Evaluation at the mention level thus
prevents the use of kappa and leaves $F_{1}$ as the only choice.

\section{Results}\label{sec:results}

% ---------------------------------------------------------------------

\subsection{Experimental protocol}\label{sec:protocol}

\noindent In \cite{Esuli:2013uq}, experiments on the UmbertoI(RadRep)
dataset were run using either \texttt{1-only} and/or \texttt{2-only}
(i.e., the portions of the data that only one coder had annotated) as
training data and \texttt{Both(1)} and/or \texttt{Both(2)} (i.e., the
portion of the data that both coders had annotated, in both versions)
as test data.

In this paper we switch the roles of training set and test set, i.e.,
use \texttt{Both(1)} or \texttt{Both(2)} as training set (since for
the purpose of this paper we need \emph{training} data with multiple,
alternative annotations) and \texttt{1-only} or \texttt{2-only} as
test set. Specifically, we run two batches of experiments, Batch1 and
Batch2. In Batch1 Coder1 plays the role of the authoritative coder
($C_{\alpha}$) and Coder2 plays the role of the non-authoritative coder
($C_{\beta}$), while in Batch2 Coder2 plays the role of $C_{\alpha}$
and Coder1 plays the role of $C_{\beta}$.

Each of the two batches of experiments is composed of:

\begin{enumerate}

\item An experiment using the authoritative setting, i.e., both training
  and test data are annotated by $C_{\alpha}$. This means training on
  \texttt{Both(1)} and testing on \texttt{1-only} (Batch1) and
  training on \texttt{Both(2)} and testing on \texttt{2-only}
  (Batch2).
 
\item An experiment using the non-authoritative setting, i.e., training
  data annotated by $C_{\beta}$ and test data annotated by
  $C_{\alpha}$. This means training on \texttt{Both(2)} and testing on
  \texttt{1-only} (Batch1) and training on \texttt{Both(1)} and
  testing on \texttt{2-only} (Batch2).

\item Experiments using the partially authoritative setting, i.e.,
  test data annotated by $C_{\alpha}$, and training data annotated in
  part by $C_{\beta}$ ($\lambda\%$ of the training documents, chosen
  at random) and in part by $C_{\alpha}$ (the remaining
  $(100-\lambda)\%$ of the training documents). We call $\lambda$ the
  \emph{corruption ratio} of the training set; $\lambda=0$ obviously
  corresponds to the fully authoritative setting while $\lambda=100$
  corresponds to the non-authoritative setting.
 
  We run experiments for each $\lambda\in\{10, 20, ..., 80, 90\}$ by
  monotonically adding, for increasing values of $\lambda$, new
  randomly chosen elements (10\% at a time) to the set of training
  documents annotated by $C_{\beta}$.
  % For increasing values of $\lambda$ we randomly select a further
  % 10\%
  % from the remaining training documents annotated by $C_{\alpha}$
  % and
  % replace them by the corresponding documents annotated by
  % $C_{\beta}$. For instance, given that the experiment with
  % $\lambda=30$
  % uses a set (let us call it $Alt_{30}$) consisting of 30\% of the
  % training documents as annotated by $C_{\beta}$, and a set (let us
  % call it $Aut_{70}$) consisting of the remaining 70\% of the
  % training documents as annotated by $C_{\alpha}$, in the experiment
  % with $\lambda=40$ we randomly select another 10\% of documents
  % from
  % $Aut_{70}$ and replace them by the corresponding documents as
  % annotated by $C_{\beta}$.
  Since the choice of training data annotated by $C_{\beta}$ is
  random, we repeat the experiment 10 times for each value of
  $\lambda\in\{10, 20, ..., 80, 90\}$, each time with a different
  random such choice.

\end{enumerate}

\noindent For each of the above train-and test experiment we compute
the intercoder agreement $\kappa(Tr,corr_{\lambda}(Tr))$ between the
non-corrupted version of the training set $Tr$ and the (partially or
fully) corrupted version $corr_{\lambda}(Tr)$ for a given value of
$\lambda$. We then take the average among the 10 values of
$\kappa(Tr,corr_{\lambda}(Tr))$ deriving from the 10 different
experiments run for a given value of $\lambda$
% (10 random choices of the elements in $Tr_{\beta}$ $\times$ 2
% batches of experiments),
and denote it as $\kappa(\lambda)$; this value indicates the average
intercoder agreement that derives by ``corrupting'' $\lambda\%$ of the
documents in the training set, i.e., by using for them the annotations
performed by the non-authoritative coder.

For each of the above train-and test experiment we also compute the
extraction accuracy (via both $F_{1}^{\mu}$ and $F_{1}^{M}$) and the
relative loss in extraction accuracy that results from the given
corruption ratio.

% $L_{F_{1}}(Tr_{\alpha},Tr_{\beta},Te_{\alpha})$ as measured on the
% test set. We then compute the mean and the variance among the 20
% values of $F_{1}(Te_{\alpha})$ and
% $L_{F_{1}}(Tr_{\alpha},Tr_{\beta},Te_{\alpha})$ deriving from the 20
% different experiments run for a given value of $\lambda$, and denote
% them as $F_{1}(x)$ and $L_{F_{1}}(x)$.

% ---------------------------------------------------------------------

\subsection{Results and discussion}\label{sec:resultsanddiscussion}

\begin{table}[t]
  \begin{center}
  \tbl{\label{tab:results}Extraction accuracy for the authoritative setting ($\lambda=0$) and non-authoritative setting ($\lambda=100$), for the LC-CRFs and HM-SVMs learners, for both batches of experiments (and for the average across the two batches), along with the resulting intercoder agreement values expressed as $\kappa(\lambda)$. Percentages indicate the loss in extraction accuracy resulting from moving from $\lambda=0$ to $\lambda=100$.}{
    \resizebox{\textwidth}{!} {
    \begin{tabular}{|c|c||c||cc|cc||cc|cc|}
      \hline
      \multicolumn{3}{|c||}{\mbox{}} & \multicolumn{4}{c||}{LC-CRFs} & \multicolumn{4}{c|}{HM-SVMs} \\
      \hline
      & $\lambda$ & $\kappa(\lambda)$ & \multicolumn{2}{c|}{$F_{1}^{\mu}$} & \multicolumn{2}{c||}{$F_{1}^{M}$} & \multicolumn{2}{c|}{$F_{1}^{\mu}$} & \multicolumn{2}{c|}{$F_{1}^{M}$} \\
      \hline\hline
      \multirow{2}{*}{Batch1} & 0 & 1.000 & 0.783 & & 0.674 & & 0.820 & & 0.693 & \\
      & 100 & 0.742 & 0.765 & (-2.35\%) & 0.668 & (-0.90\%) & 0.786 & (-4.33\%) & 0.688 & (-0.73\%) \\
      \hline
      \multirow{2}{*}{Batch2} & 0 & 1.000 & 0.808 & & 0.752 & & 0.817 & & 0.754 & \\
      & 100 & 0.742 & 0.733 & (-10.23\%) & 0.654 & (-14.98\%) & 0.733 & (-11.46\%) & 0.625 & (-20.64\%) \\
      \hline
      \hline
      \multirow{2}{*}{Average} & 0 & 1.000 & 0.795 & & 0.713 & & 0.819 & & 0.724 & \\
      & 100 & 0.742 & 0.749 & (-6.14\%) & 0.661 & (-7.87\%) & 0.760 & (-7.76\%) & 0.657 & (-10.20\%) \\
      \hline
    \end{tabular}
    }}
  \end{center}
\end{table}

\begin{figure}[t]
  \begin{center}
    \caption{Microaveraged $F_{1}$ (left) and macroaveraged $F_{1}$
    (right) as a function of the fraction $\lambda$ of the training
    set that is annotated by $C_{\beta}$ instead of $C_{\alpha}$
    (``corruption ratio''), using \textbf{LC-CRFs} (top) and \textbf{HM-SVMs} (bottom) as learning
    algorithms. \blue{The dashed line represents the experiments in
    Batch1, the dotted line represents those in Batch2, and the solid
    one represents the average between the two batches. The vertical
    bars indicate, for each $\lambda\in\{10, 20, ..., 80, 90\}$, the
    standard deviation across the 10 runs deriving from the 10 random
    choices of $corr_{\lambda}(Tr)$.}}    
    \scalebox{.33}[.33]{\includegraphics{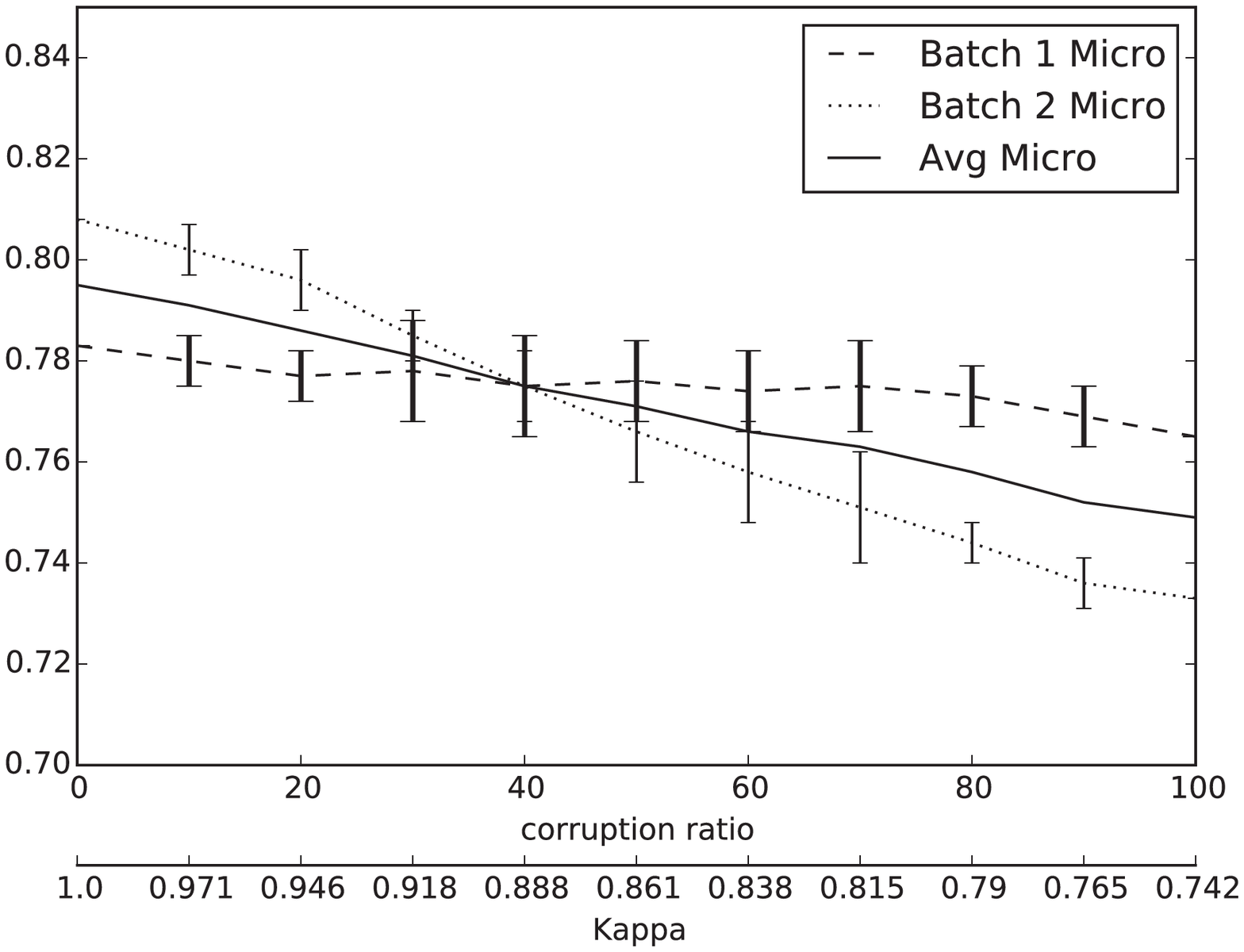}}
    \scalebox{.33}[.33]{\includegraphics{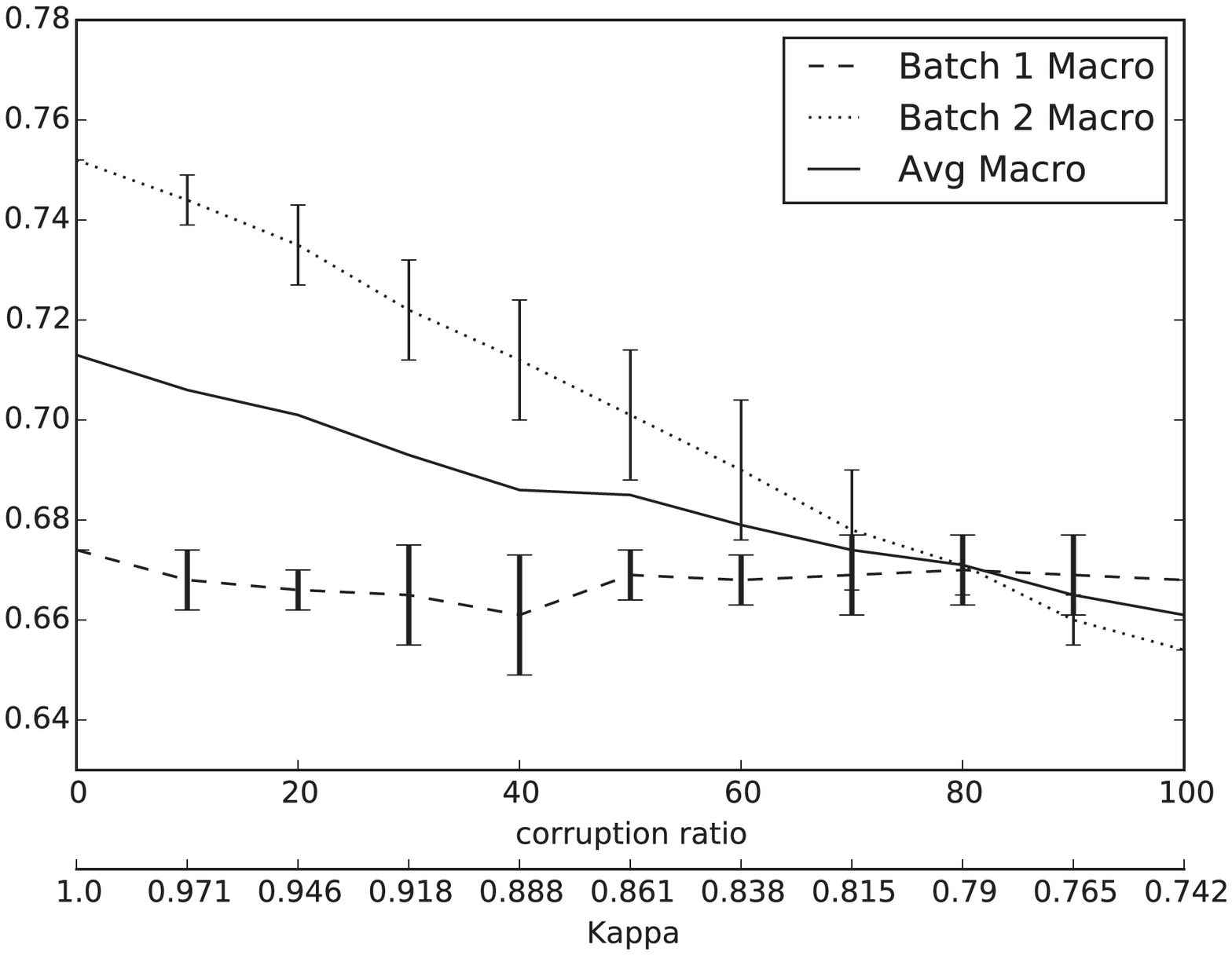}}
    \scalebox{.33}[.33]{\includegraphics{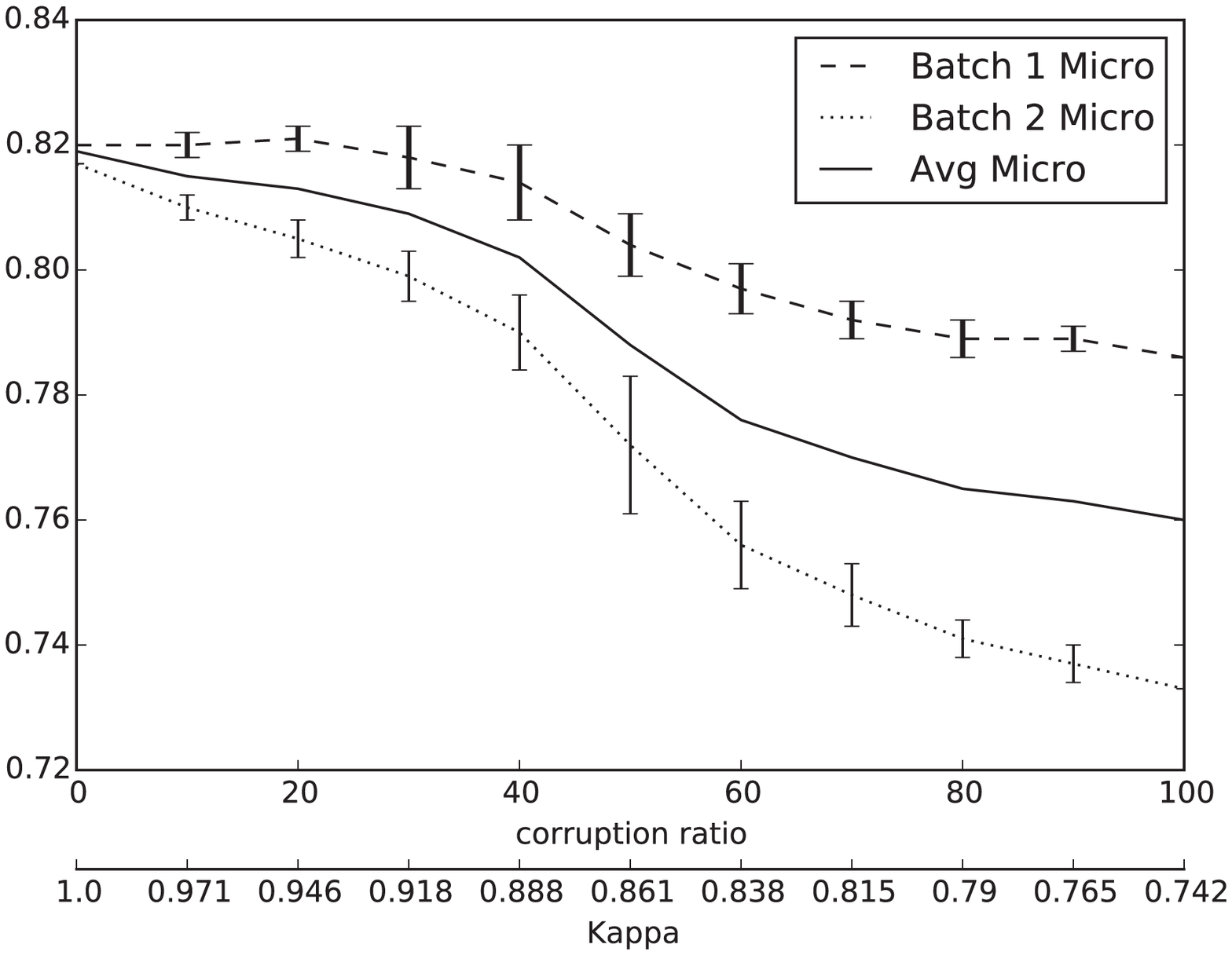}}
    \scalebox{.33}[.33]{\includegraphics{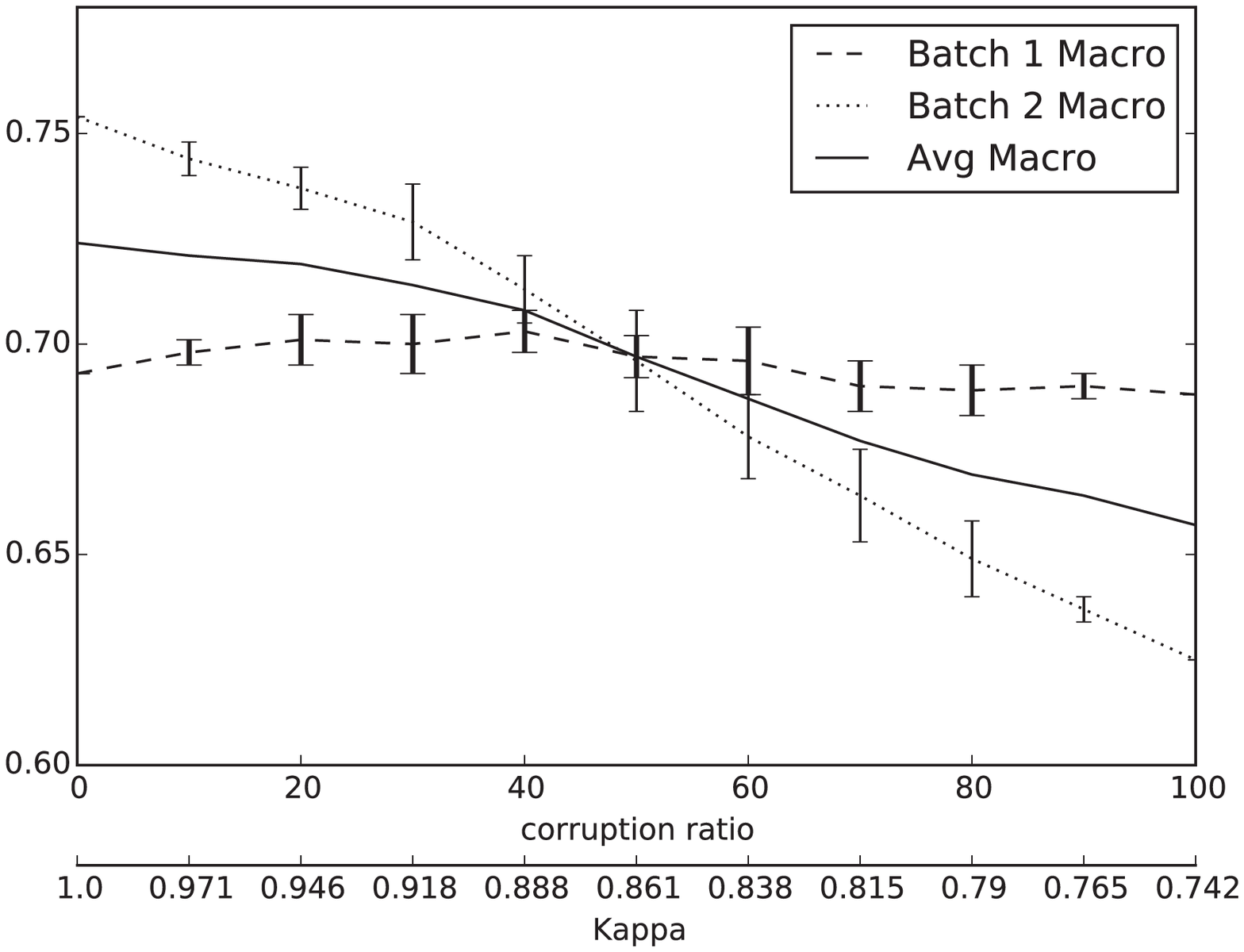}}
    \label{fig:F1CRFsandSVMs}
  \end{center}
\end{figure}

\begin{figure}[t]
  \begin{center}
    \caption{\blue{Microaveraged (left) and macroaveraged (right) precision
    (top) and recall (bottom) as a function of the fraction $\lambda$
    of the training set that is annotated by $C_{\beta}$ instead of
    $C_{\alpha}$ (``corruption ratio''), using \textbf{LC-CRFs} as a
    learning algorithm.}}
    \scalebox{.33}[.33]{\includegraphics{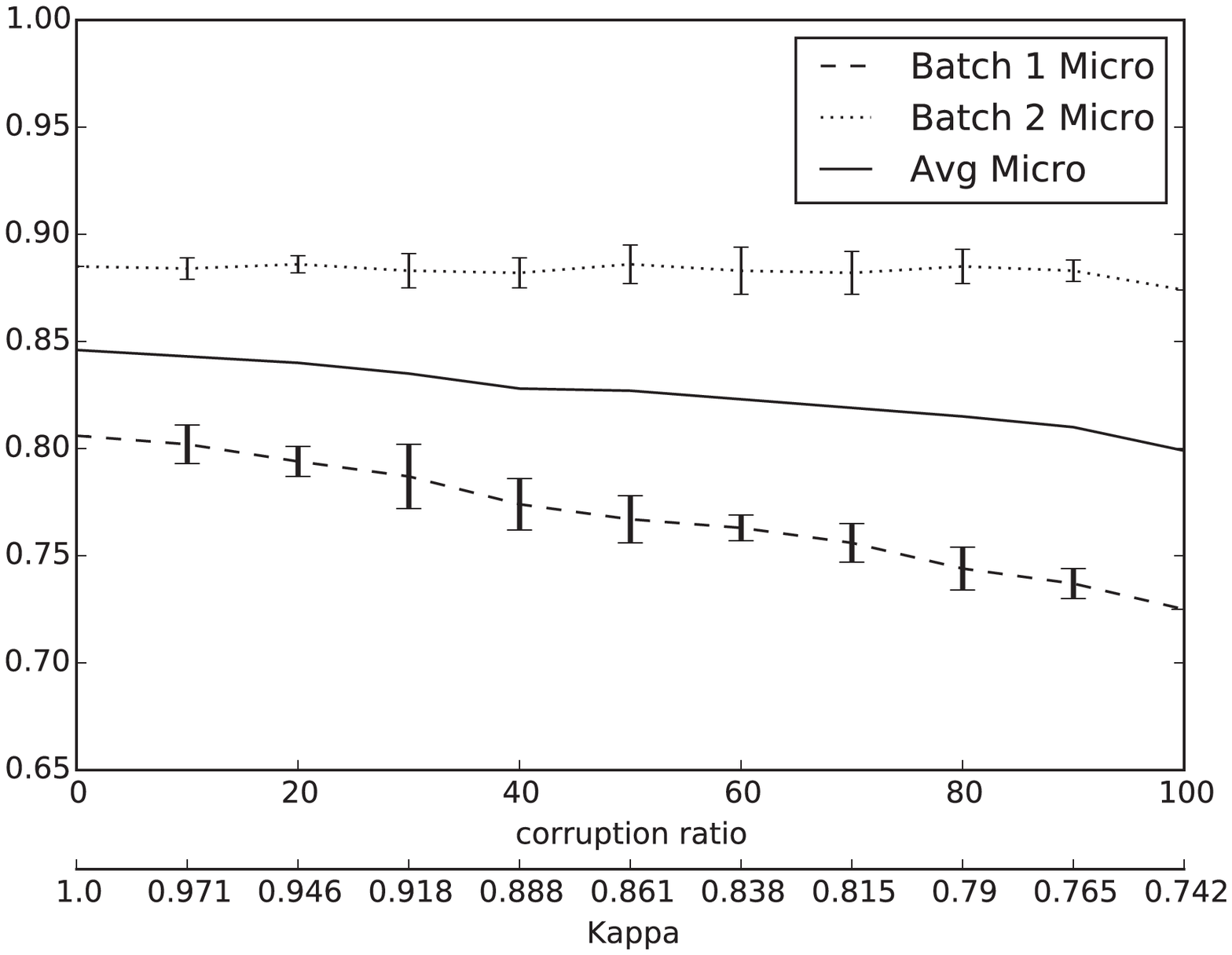}}
    \scalebox{.33}[.33]{\includegraphics{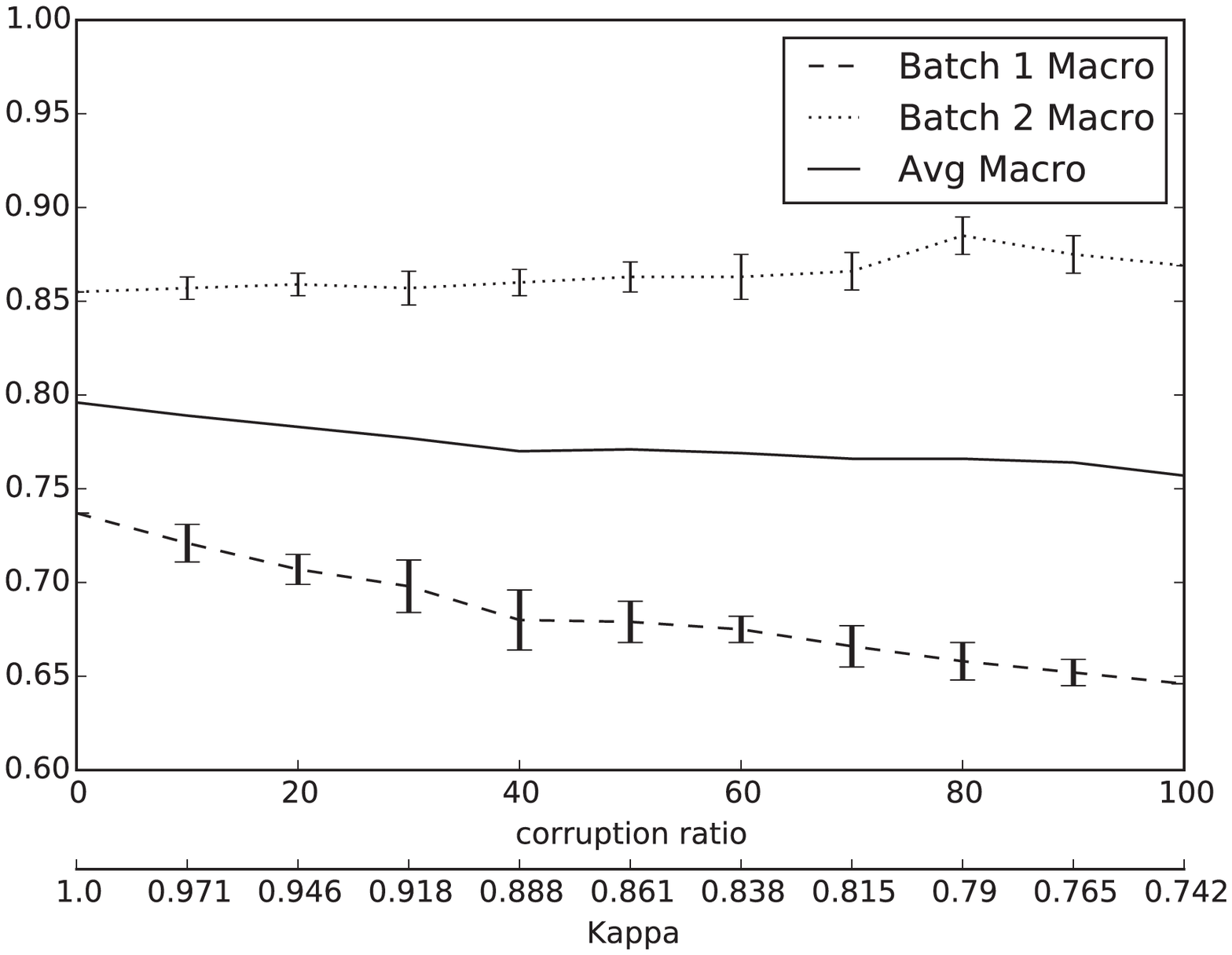}}
    \scalebox{.33}[.33]{\includegraphics{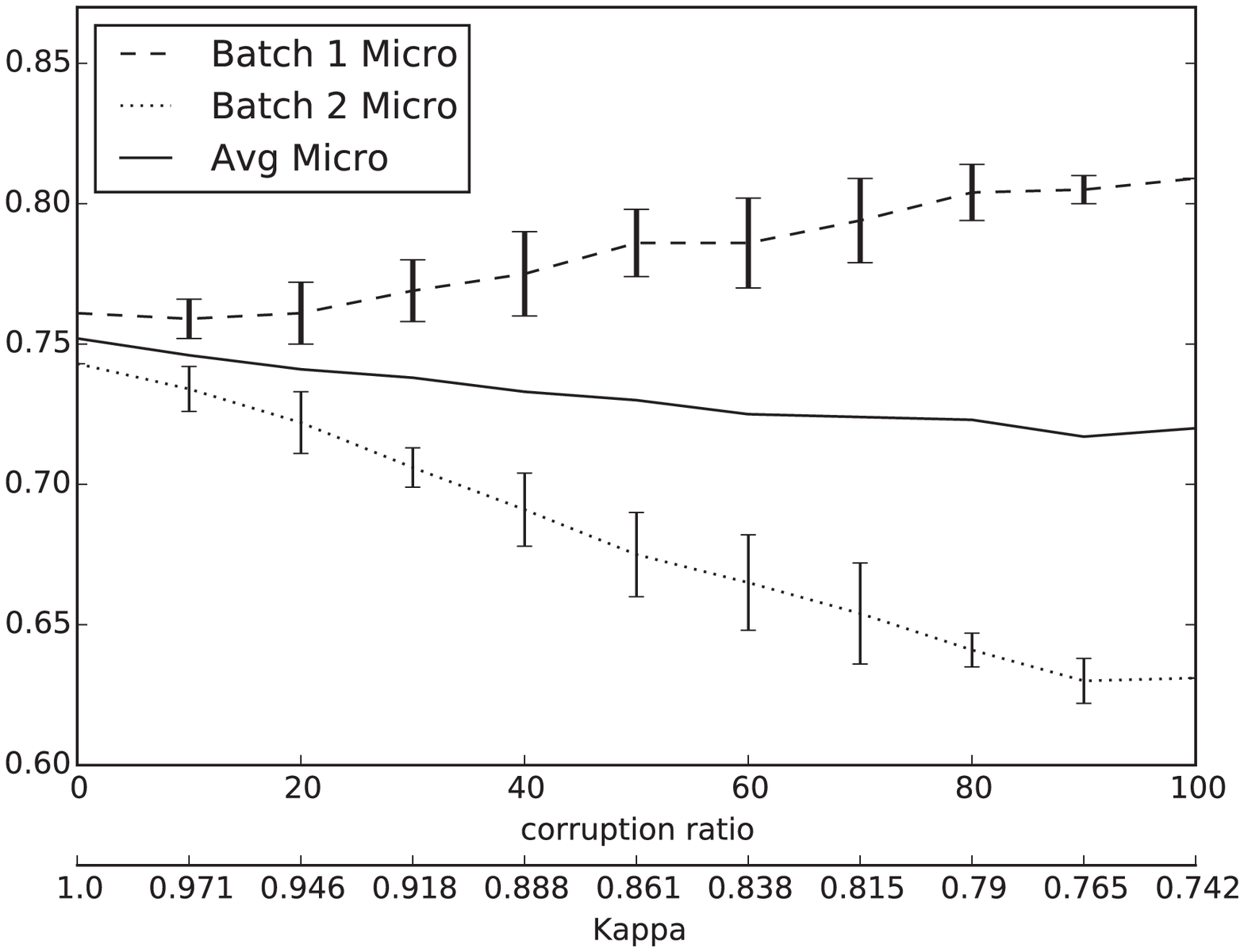}}
    \scalebox{.33}[.33]{\includegraphics{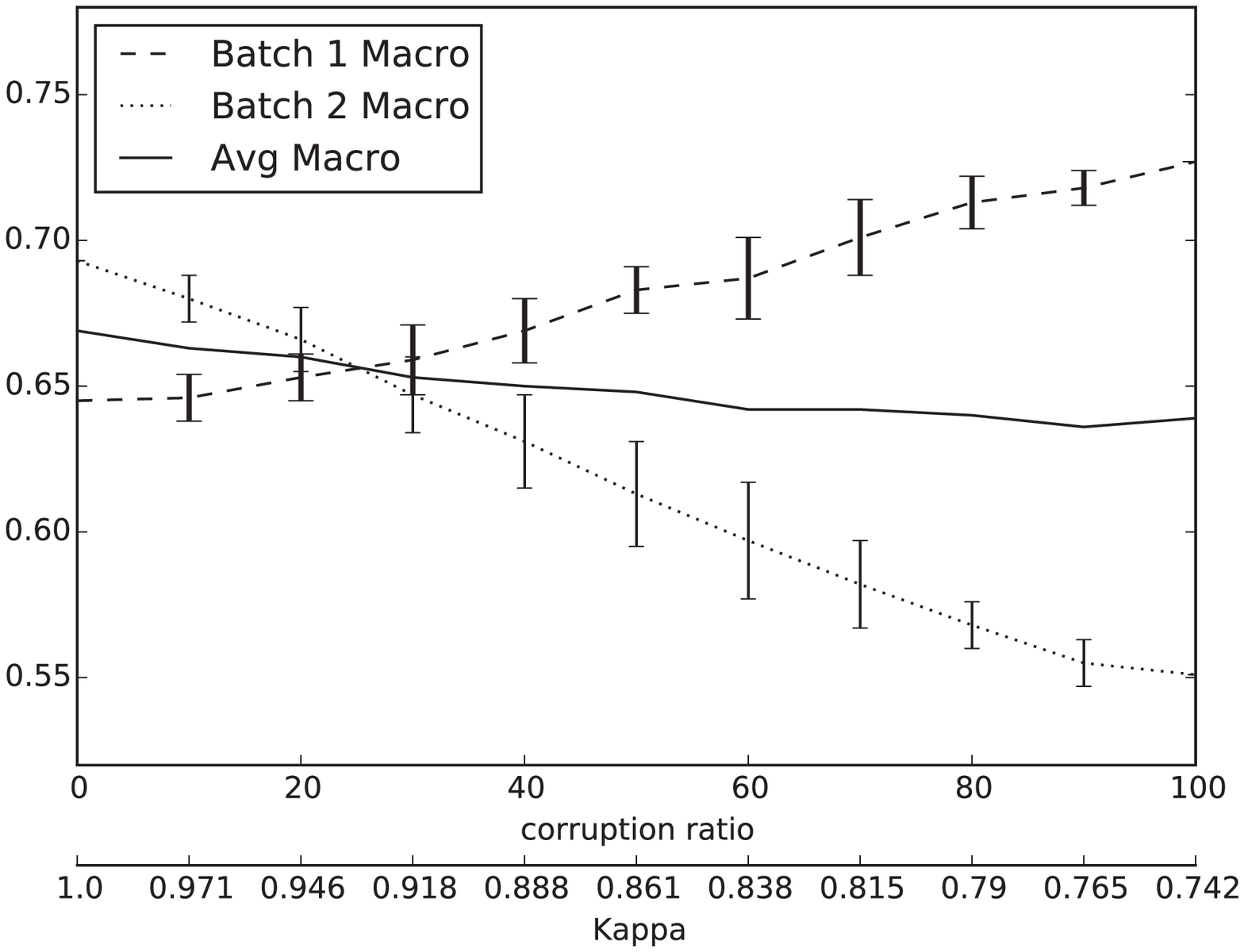}}
    \label{fig:precisionandrecallCRF}
  \end{center}
\end{figure}

\begin{figure}[t]
  \begin{center}
    \caption{\blue{Microaveraged (left) and macroaveraged (right) precision
    (top) and recall (bottom) as a function of the fraction $\lambda$
    of the training set that is annotated by $C_{\beta}$ instead of
    $C_{\alpha}$ (``corruption ratio''), using \textbf{HM-SVMs} as a
    learning algorithm.}}
    \scalebox{.33}[.33]{\includegraphics{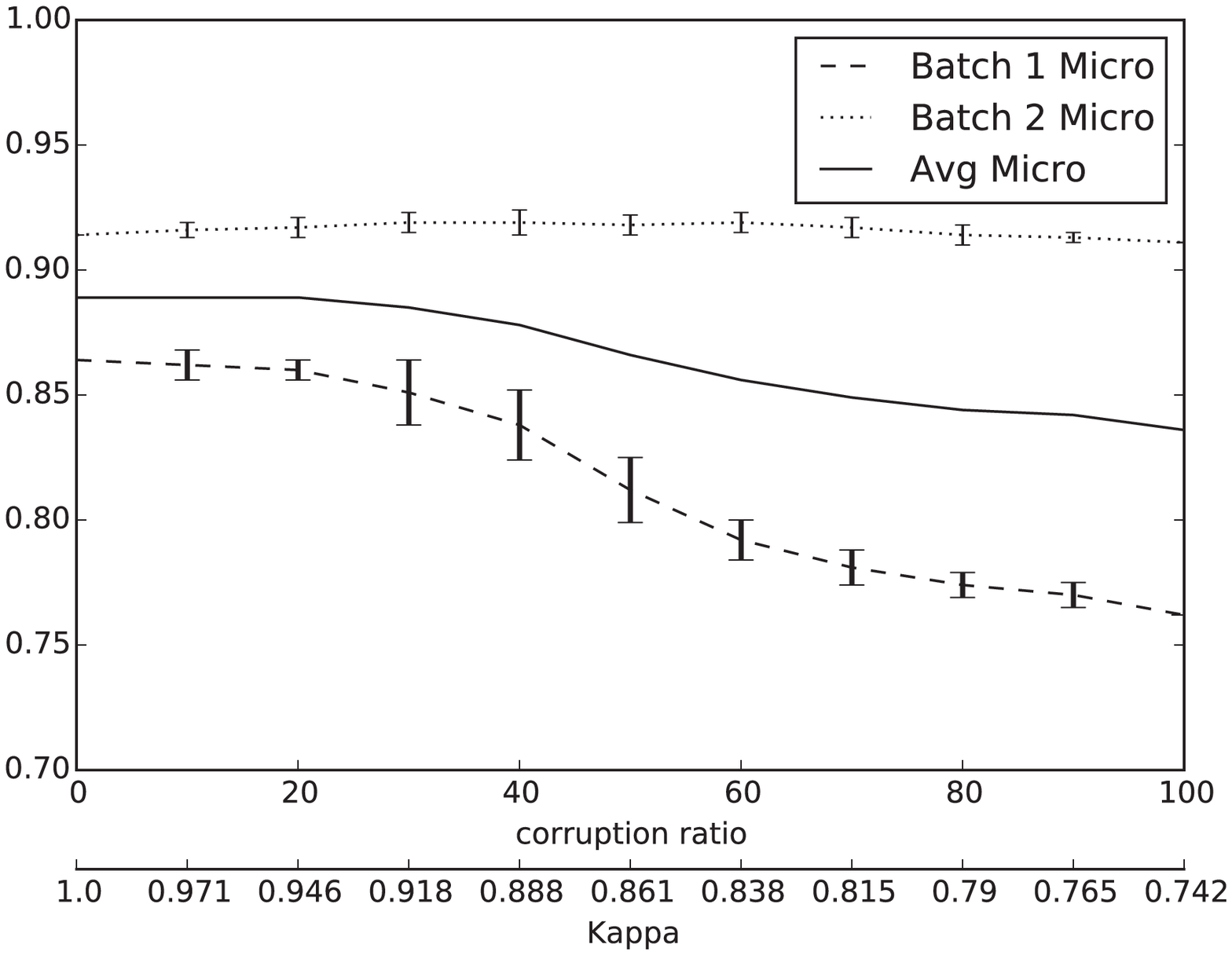}}
    \scalebox{.33}[.33]{\includegraphics{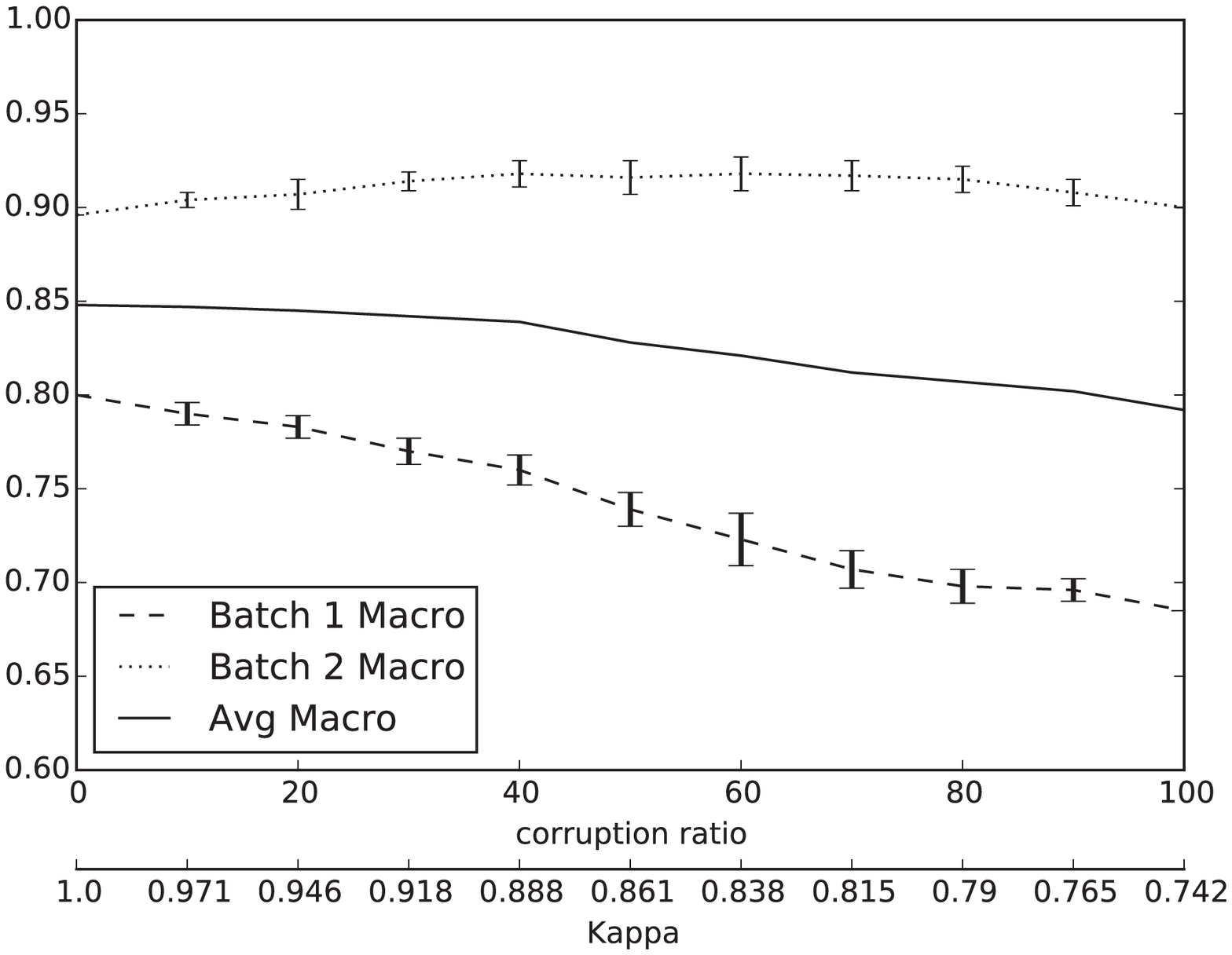}}
    \scalebox{.33}[.33]{\includegraphics{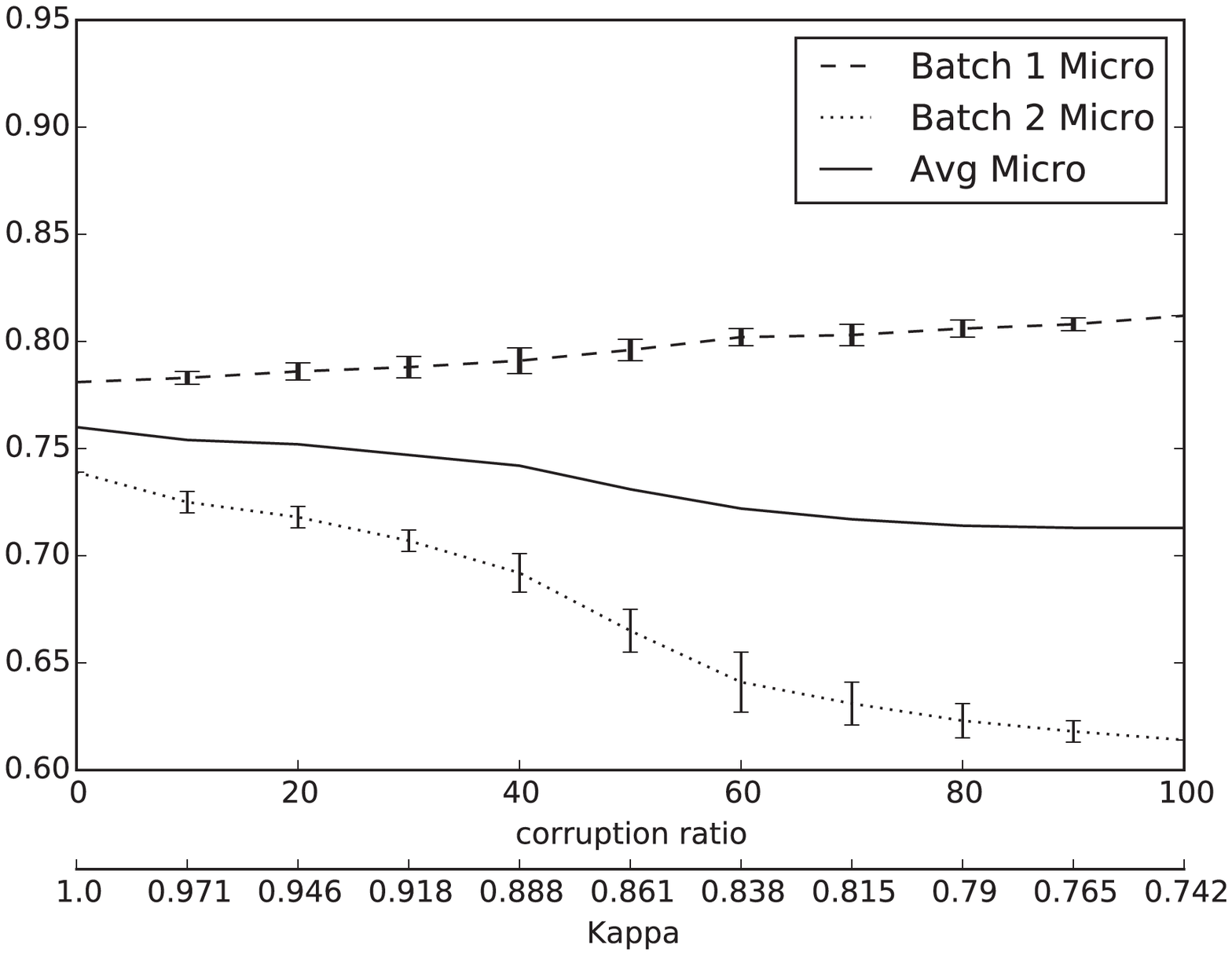}}
    \scalebox{.33}[.33]{\includegraphics{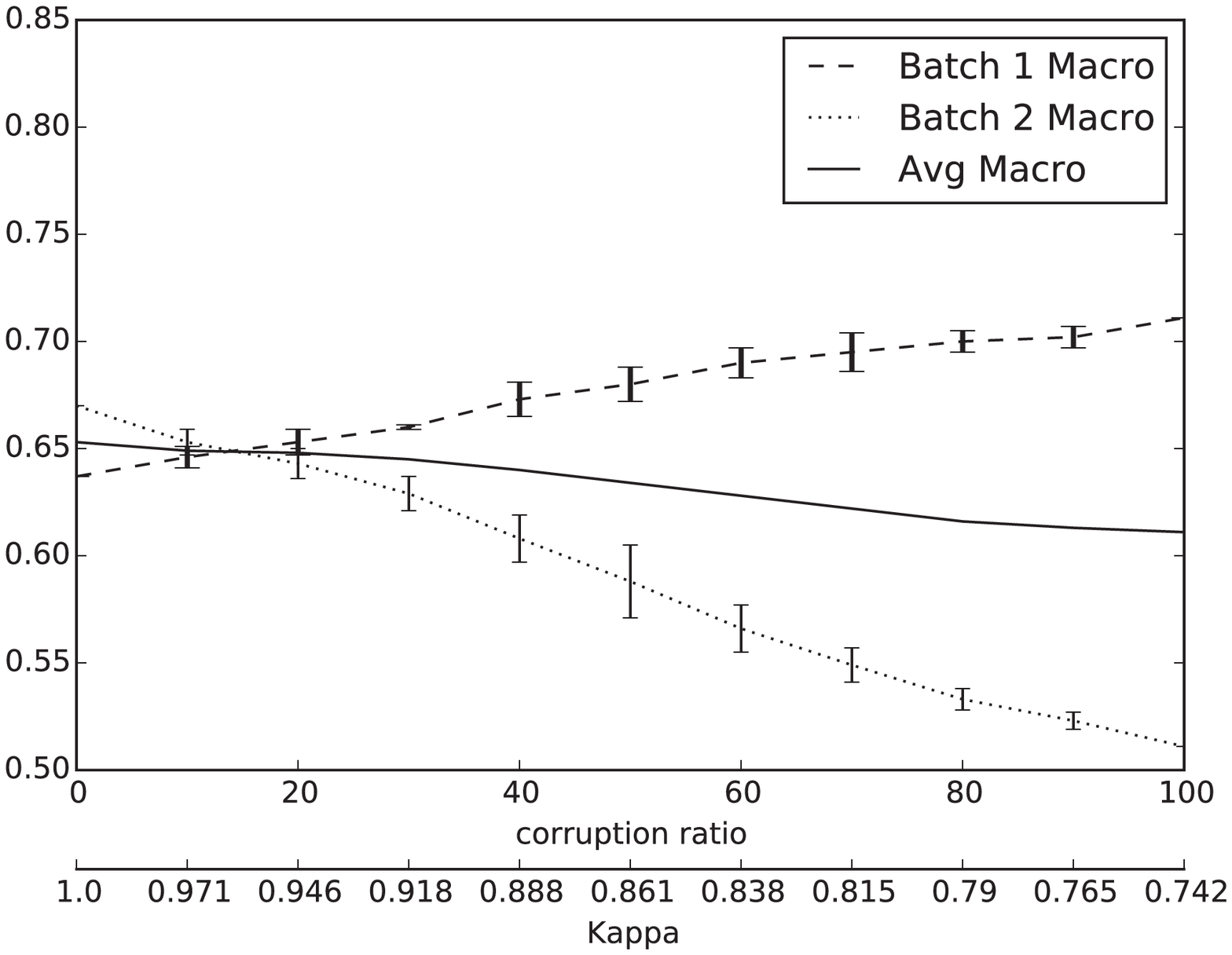}}
    \label{fig:precisionandrecallSVM}
  \end{center}
\end{figure}

\noindent Table \ref{tab:results} reports extraction accuracy figures
for the authoritative and non-authoritative settings, for both learners,
both batches of experiments, and along with the resulting intercoder
agreement values. Figure \ref{fig:F1CRFsandSVMs}
illustrates the results of our experiments by plotting $F_{1}$ as a
function of the corruption ratio $\lambda$, using LC-CRFs \blue{and
HM-SVMs} as the learning algorithm, respectively; for each value of
$\lambda$, the corresponding level of interannotator agreement
$\kappa(\lambda)$ (as averaged across the two batches) is also
indicated. \blue{Figure \ref{fig:precisionandrecallCRF} plots instead
precision and recall as a function of $\lambda$ for the LC-CRFs
experiments, while Figure \ref{fig:precisionandrecallSVM} does the
same for the HM-SVMs experiments.}

% ---------------------------------------------------------------------

\subsubsection{Macroaveraged values are lower than microaveraged
ones}\label{sec:microbetterthanmacro}

\noindent A first fact to be observed is that macroaveraged
($F_{1}^{M}$) results are always lower than the corresponding
microaveraged ($F_{1}^{\mu}$) results. This is unsurprising, and
conforms to a well-known pattern. In fact, microaveraged effectiveness
scores are heavily influenced by the accuracy obtained on the concepts
most frequent in the test set (i.e., on the ones that label many test
\blue{t-units}); for these concepts accuracy tends to be higher, since these
concepts also tend to be more frequent in the training
set. Conversely, in macroaveraged effectiveness measures, each concept
counts the same, which means that the low-frequency concepts (which
tend to be the low-performing ones too) have as much of an impact as
the high-frequency ones. See \cite[pp.\ 591--593]{Debole05} for a
thorough discussion of this point in a text classification context.

% ---------------------------------------------------------------------

\subsubsection{\blue{HM-SVMs outperform
LC-CRFs}}\label{sec:SVMsoutperformCRFs}

\noindent \blue{A second fact that emerges is that HM-SVMs outperform
LC-CRFs, on both batches, both settings (authoritative and
non-authoritative), and both evaluation measures ($F_{1}^{\mu}$ and
$F_{1}^{M}$); e.g., on the authoritative setting, and as an average
across the two batches, HM-SVMs obtain $F_{1}^{\mu}=0.819$ (while
LC-CRFs obtain 0.795) and $F_{1}^{M}=0.724$ (while LC-CRFs obtain
0.713). Aside from their different levels of effectiveness, the two
learners behave in a qualitatively similar way as a function of
$\lambda$, as evident from a comparison of Figures
\ref{fig:precisionandrecallCRF} and
\ref{fig:precisionandrecallSVM}. However, we will not dwell on this
fact any further since the relative performance of the learning
algorithms is not the main focus of the present study; as will be
evident in the discussion that follows, most insights obtained from
the LC-CRFs experiments are qualitatively confirmed by the HM-SVMs
experiments, and vice versa.}

% ---------------------------------------------------------------------

\subsubsection{Coder1 generates less accuracy than
Coder2}\label{sec:Coder1betterthanCoder2}

\noindent A third fact that may be noted (from Table
\ref{tab:results}) is that, for $\lambda=0$, there is a substantive
difference in accuracy values between the two coders, with Coder2
usually generating higher accuracy than Coder1.
% \diescomment{Questa cosa non è più vera nei nuovi esperimenti,
% escluso la $F_1^\mu$ ottenuta con le SVM dove l'accuracy del Batch 1
% è sempre superiore.}
This fact can be especially appreciated at the macroaveraged level
(where for LC-CRFs we have $F_{1}^{M}=0.674$ for Coder1 and
$F_{1}^{M}=0.752$ for Coder2, and for HM-SVMs we have
$F_{1}^{M}=0.693$ for Coder1 and $F_{1}^{M}=0.754$ for Coder2), while
the difference is less clearcut at the microaveraged level (where for
LC-CRFs we have $F_{1}^{M}=.0.783$ for Coder1 and $F_{1}^{M}=0.808$
for Coder2, and for HM-SVMs we have $F_{1}^{M}=0.820$ for Coder1 and
$F_{1}^{M}=0.817$ for Coder2); this indicates that the codes where
Coder2 especially shines are the low-frequency ones.

In principle, there might be several reasons for this difference in
accuracy values between the two coders. The documents in
\texttt{2-only} might be ``easier'' to code automatically than those
in \texttt{1-only}; or the distributions of \texttt{Both(1)} and
\texttt{1-only} might be less similar to each other than the
distributions of \texttt{Both(2)} and \texttt{2-only}, thus verifying
the i.i.d.\ assumption to a higher degree; or Coder2 might simply be
more self-consistent in her annotation style than Coder1.

In order to check whether the last of these three hypotheses is true
% One fact that might explain this difference is a different level of
% self-consistency in the labelling work of the two coders. We
% conjecture that this difference in behaviour between Batch1 and
% Batch2 may be due to the fact that Coder1 is more
% ``self-consistent'' than Coder2.
we have performed four $k$-fold cross-validation ($k$-FCV) experiments
(for \texttt{Both(1)} and \texttt{Both(2)}, and for LC-CRFs and
HM-SVMs, in all combinations), using $k=20$. Intuitively, a higher
accuracy value resulting from a $k$-FCV test means a higher level of
self-consistency, since if the same coding style is consistently used
to label a dataset, a system tends to encounter in the testing phase
the same labelling patterns it has encountered in the training phase,
which is conducive to higher accuracy. Of course, the results of such
a test are difficult to interpret if the goal is to assess the
self-consistency of a coder in \emph{absolute} terms (since we do not
know what values of $F_{1}$ correspond to what levels of
self-consistency), but they are not if the goal is simply to establish
which of the two is the more self-consistent, since the two
experiments are run on the same documents.  The results of our two
$k$-FCV experiments are reported in Table
\ref{tab:crossvalidation}. From this table we can see that the
accuracy on \texttt{Both(2)} is substantially higher than the one
obtained on \texttt{Both(1)}, thus indicating that Coder2 is indeed
more self-consistent than Coder1. This is thus the likely explanation
of the higher levels of accuracy obtained on the dataset annotated,
for both training and test, by Coder2.

\begin{table}[t]
  \begin{center}
  \tbl{\label{tab:crossvalidation}Results of the 20-fold cross-validation tests on \texttt{Both(1)} and \texttt{Both(2)}, for LC-CRFs \blue{and HM-SVMs}.}{
    \begin{tabular}{|c||c|c||c|c|}
      \hline
      \mbox{} & \multicolumn{2}{c||}{LC-CRFs} & \multicolumn{2}{c|}{HM-SVMs} \\
      \cline{2-5}
      & $F_{1}^{\mu}$ & $F_{1}^{M}$ & $F_{1}^{\mu}$ & $F_{1}^{M}$ \\
      \hline\hline
      \multirow{1}{*}{\texttt{Both(1)}} & 0.829 & 0.735 & 0.842 & 0.737 \\
      \hline
      \multirow{1}{*}{\texttt{Both(2)}} & 0.838 & 0.771 & 0.850 & 0.787\\
      \hline
    \end{tabular}
    }
  \end{center}
\end{table}

% ---------------------------------------------------------------------

\subsubsection{\blue{Overannotation and
underannotation}}\label{sec:Batch2betterthanBatch1}

\noindent \blue{A fourth, even more interesting fact we may observe is
that accuracy as a function of the corruption ratio varies much less
for Batch1 than for Batch2, since for this latter we witness a much
more substantial drop in going from $\lambda=0$ to $\lambda=100$. We
conjecture that this may be due to the different annotation style of
the two coders; the rest of this subsection will be devoted to
explaining the rationale of this conjecture. }

\blue{As evident from Table \ref{tab:annotationstats}, Coder2
annotates, as instances of the concepts of interest, more mentions
(+15.7\%) and also more tokens per mention (+15.6\%) than Coder1;
relatively to each other, Coder1 is thus an \emph{underannotator}
while Coder2 is an \emph{overannotator}. Since, as noted in Section
\ref{sec:introduction}, learning algorithms learn to replicate the
subjective annotation style of their supervisors, a system trained on
data annotated by an overannotator will itself tend to overannotate;
conversely, a system trained by an underannotator will itself tend to
underannotate.}

\blue{Overannotation results in more true positives and more false
positives. The plots in Figures \ref{fig:precisionandrecallCRF} and
\ref{fig:precisionandrecallSVM} show that when, as a consequence of
increased values of $\lambda$, the number of training documents
annotated by an overannotator increases (as is the case of Batch1),
precision suffers somehow (due to the fact that, along with more true
positives, there are also more false positives), but this is
compensated by an increase in recall (due to an increased number of
true positives); as a result, as shown in Figure \ref{fig:F1CRFsandSVMs} (and in Table \ref{tab:results} too), the drop in
$F_{1}$ resulting from moving to $\lambda=0$ to $\lambda=100$ is very
limited.  Figures \ref{fig:precisionandrecallCRF} and
\ref{fig:precisionandrecallSVM} instead show that when, as a
consequence of increased values of $\lambda$, the number of training
documents annotated by an underannotator increases (as is the case for
Batch2), recall drops substantially (due to the decreased number of
true positives), and this drop is not compensated by the stability of
precision (which is due to the combined effect of a decrease in true
positives and a decrease in false positives); as a result, as shown in
Figure \ref{fig:F1CRFsandSVMs} (see also Table
\ref{tab:results}), the drop in $F_{1}$ resulting from moving to
$\lambda=0$ to $\lambda=100$ is much more substantial than for
Batch1.}

In order to check whether the decreases in accuracy between the
$\lambda=0$ and the $\lambda=100$ settings is statistically
significant we have performed an \emph{approximate randomization test}
(ART) \cite{Chinchor93}.
% , which determines if there is a statistically significant
% difference between the data automatically annotated at $\lambda=0$
% corruption ratio and those annotated at $\lambda=100$ corruption
% ratio.
In this test the difference is considered statistically significant if
the resulting $p$ value is $<0.05$. Two advantages of the ART are that

\begin{enumerate}

\item unlike the t-test, the ART does not require the data to be
  normally distributed;

\item unlike the Wilcoxon signed-rank test, the ART can be applied to
  multivariate non-linear evaluation measures such as $F_{1}$
  \cite{yeh2000more}.

\end{enumerate}

% \fabscomment{Il Wilcoxon signed-rank test qui citato e il Wilcoxon
% rank-sum test che avevi usato precedentemente (vedi la parte
% commentata che avevamo scritto) sono due test differenti, almeno
% secondo Wikipedia; qui intendevi proprio dire ``Wilcoxon signed-rank
% test''?} \diescomment{Si, quello che avevo usato in precendenza era
% quello che ti dicevo essere errato completamente. Il test giusto da
% citare è Wilcoxon signed-rank test.} \fabscomment{Qui bisogna essere
% più precisi: cosa vuol dire ``sophisticated''? Vuol dire non-linear?
% multivariate?} \diescomment{vuol dire non-linear}

\noindent The results of our statistical significance tests are
reported in Table \ref{tab:statsign}. \blue{These results essentially
confirm the observations above, i.e., that \emph{in Batch2 the drop in
performance resulting from having the training set annotated by the
non-authoritative coder (instead of the authoritative one) is not
statistically significant}, while (with the exception of the
$F_{1}^{\mu}$ results for HM-SVMs) it is statistically significant for
Batch1.}

\begin{table}[t]
  \begin{center}
  \tbl{\label{tab:statsign}Results of the approximate randomization test, measuring the statistical significance of the difference between the accuracy of the system trained at $\lambda=0$ and the accuracy of the system trained at $\lambda=100$. Results are reported for both learners (LC-CRFs \blue{and HM-SVMs}), both batches, and both evaluation measures ($F_{1}^{\mu}$ and $F_{1}^{M}$).}{
     % \resizebox{\textwidth}{!} {
    \begin{tabular}{|c||c|c||c|c|}
      \hline
      \mbox{} & \multicolumn{2}{c||}{LC-CRFs} & \multicolumn{2}{c|}{HM-SVMs} \\
      \cline{2-5}
      & $F_{1}^{\mu}$ & $F_{1}^{M}$ & $F_{1}^{\mu}$ & $F_{1}^{M}$ \\
      \hline\hline
      Batch1 & 0.0859 & 0.6207 & 0.0001 & 0.5040 \\
      \hline
      Batch2 & 0.0001 & 0.0001 & 0.0001 & 0.0001 \\
      \hline
    \end{tabular}
     }
  \end{center}
\end{table}

% ---------------------------------------------------------------------

\subsubsection{\blue{Caveats}}\label{sec:discussion}

\noindent The experiments discussed in this paper do not allow us to reach hard conclusions about the robustness of information extraction systems to imperfect training data quality, for several reasons:

\begin{enumerate}

 \item The results obtained should be confirmed by additional experiments carried out on other datasets; unfortunately, as noted in Footnote \ref{foot:datasets}, we were not able to locate any other publicly available dataset with the required characteristics (that is, containing at least some doubly annotated documents).
 
 \item The dataset used here is representative of only a specific type of imperfect training data quality, i.e., the one deriving from the fact that the training data were annotated by a coder different (albeit equally expert) from the one who annotated the test set. Other types do exist, however, as noted in the introduction.
 
 \item Even the results reported here are somehow contradictory, since a statistically significant drop in performance was observed in Batch1 while no such statistically significant drop was observed in Batch2.

\end{enumerate}

\noindent However, one interesting fact that has emerged from the
present study (and that will need to be confirmed by additional
experiments, should other datasets become available) is that, as
argued in detail in Section \ref{sec:Batch2betterthanBatch1}, the lack
of a statistically significant drop in performance observed in Batch2
seems to be due to the fact that the non-authoritative coder who
annotated the training set had an \emph{overannotating}
behaviour. This \emph{might} suggest (emphasis meaning that prudence
should be exercised) that, should there be a need for having a
training set annotated by someone different from the authoritative
coder, underannotation should be discouraged much more than
overannotation.

\section{Conclusions}\label{sec:conclusions}

\noindent Few researchers have investigated the loss in accuracy that
occurs when a supervised learning algorithm is fed with training data
of suboptimal quality. We have done this for the first time in the
case of information extraction systems (trained via supervised
learning) as applied to the detection of mentions of concepts of
interest in medical notes. Specifically, we have tested to what extent
extraction accuracy suffers when the person who has annotated the test
data (the ``authoritative coder''), whom we must assume to be the
person to whose judgment we conform irrespectively of her level of
expertise, is different from the person who has labelled the training
data (the ``non-authoritative coder''). \blue{Our experimental
results, that we have obtained on a dataset of 500 mammography reports
annotated according to 9 concepts of interest, are somehow surprising,
since they indicate that the resulting drop in accuracy is not always
statistically significant. In our experiments, no statistically
significant drop was observed when the non-authoritative coder had a
tendency to overannotate, while a substantial, statistically
significant drop was observed when the non-authoritative coder was an
underannotator; however, experiments on more doubly (or even multiply) annotated datasets will be needed
to confirm or disconfirm these initial findings.}  Since labelling
cost is an important issue in the generation of training data (with
senior coders costing much more than junior ones, and with internal
coders costing much more than ``mechanical turkers''), results of this
kind may give important indications as to the cost-effectiveness of
low-cost annotation work.

This paper is a first attempt to investigate the impact of
less-than-sterling training data quality on the accuracy of medical
concept extraction systems, and more work is needed to validate the
conjectures that we have made based on our experimental results. As
repeatedly mentioned in this paper, one limit of the present work is
the fact that only one dataset was used for the experiments. This was
due to the unfortunate lack of other publicly available medical
datasets that contain (at least a subset of) textual records
independently labelled by two different coders; datasets with these
characteristics have been used in the past in published research but
are not made available to the rest of the scientific community. We
hope that the increasing importance of text mining applications in
clinical practice, and the importance of shared datasets for fostering
advances in this field, will generate a new kind of awareness on the
need to make more datasets available to the scientific community.

\bibliographystyle{ACM-Reference-Format-Journals}
 \bibliography{RecentStuff}
%\bibliography{RecentStuff}

% ------------------------------------------------------------------------
\end{document}